\documentclass[letterpaper]{article} 
\usepackage{aaai2026}  
\usepackage{times}  
\usepackage{helvet}  
\usepackage{courier}  
\usepackage[hyphens]{url}  
\usepackage{graphicx} 
\usepackage{natbib}  
\usepackage{caption} 
\usepackage{amsfonts} 
\usepackage{amssymb} 
\usepackage{algorithm}

\usepackage{soul} 
\usepackage{comment} 
\usepackage{witharrows}
\newcommand{\fab}[1]{{\color{blue} #1}}

\usepackage{caption}
\usepackage{multirow}
\usepackage{subcaption}
\usepackage{amsmath}
\usepackage{cleveref}
\usepackage{algpseudocode}
\usepackage{booktabs}
\usepackage[acronym]{glossaries}
\newacronym {tinyml}{TinyML}{Tiny Machine Learning}
\newacronym {ml}{ML}{Machine Learning}
\newacronym {nn}{NN}{Neural Network}
\newacronym {od}{OD}{Object Detection}
\newacronym{odl}{ODL}{On-Device Learning}
\newacronym{nms}{NMS}{Non-Maximum Suppression}
\newacronym {sota}{SotA}{State-of-the-Art}
\newacronym{fcnn}{FCNN}{Fully Connected Neural Network}
\newacronym {cnn}{CNN}{Convolutional Neural Network}
\newacronym {dnn}{DNN}{Deep Neural Network}
\newacronym{knn}{KNN}{K-nearest neighbors}
\newacronym {cat-forg}{CF}{Catastrophic Forgetting}
\newacronym{mcu}{MCU}{Microcontroller Unit}
\newacronym{voc}{VOC}{PASCAL Visual Object Classes}
\newacronym{coco}{COCO}{Common Objects in Context}
\newacronym{yolo}{YOLO}{You Only Look Once}
\newacronym{iou}{IoU}{Intersection over Union}
\newacronym{map}{mAP}{mean Average Precision}
\newacronym{ai}{AI}{Artificial Intelligence}

\newacronym{continual-learning}{CL}{Continual Learning}
\newacronym{qat}{QAT}{Quantization Aware Training}
\newacronym{ptq}{PTQ}{Post-Training Quantization}
\newacronym{ace}{ACE}{Arithmetic Computation Effort}
\newacronym{mac}{MAC}{Multiply-Accumulate}
\newacronym{sgd}{SGD}{Stochastic Gradient Descent}
\newacronym{ema}{EMA}{exponential moving average}
\newacronym{pact}{PACT}{PArameterized Clipping acTivation}

\usepackage{algorithm}
\usepackage{algpseudocode}
\algnewcommand\algorithmicinput{\textbf{Input:}}
\algnewcommand\algorithmicoutput{\textbf{Output:}}
\algnewcommand\Input{\item[\algorithmicinput]}
\algnewcommand\Output{\item[\algorithmicoutput]}
\usepackage{soul}

\usepackage{newfloat}
\usepackage{listings}
\DeclareCaptionStyle{ruled}{labelfont=normalfont,labelsep=colon,strut=off} 
\floatstyle{ruled}
\newfloat{listing}{tb}{lst}{}
\floatname{listing}{Listing}
\nocopyright 

\title{DQT: Dynamic Quantization Training \\via Dequantization-Free Nested Integer Arithmetic}
\author {
    Hazem Hesham Yousef Shalby\textsuperscript{\rm 1},
    Fabrizio Pittorino\textsuperscript{\rm 1},
    Francesca Palermo\textsuperscript{\rm 2},
    Diana Trojaniello\textsuperscript{\rm 2},
    Manuel Roveri\textsuperscript{\rm 1}
}
\affiliations {
    \textsuperscript{\rm 1}Politecnico di Milano, Milan, Italy \quad
    \textsuperscript{\rm 2}EssilorLuxottica Smart Eyewear Lab, Milan, Italy\\
    hazemhesham.shalby@polimi.it
}

\usepackage{bibentry}

\begin{document}

\maketitle

\begin{abstract}
The deployment of deep neural networks on resource-constrained devices relies on quantization. While static, uniform quantization applies a fixed bit-width to all inputs, it fails to adapt to their varying complexity. Dynamic, instance-based mixed-precision quantization promises a superior accuracy-efficiency trade-off by allocating higher precision only when needed. However, a critical bottleneck remains: existing methods require a costly dequantize-to-float and requantize-to-integer cycle to change precision, breaking the integer-only hardware paradigm and compromising performance gains. This paper introduces Dynamic Quantization Training (DQT), a novel framework that removes this bottleneck. At the core of DQT is a nested integer representation where lower-precision values are bit-wise embedded within higher-precision ones. This design, coupled with custom integer-only arithmetic, allows for on-the-fly bit-width switching through a near-zero-cost bit-shift operation.  This makes DQT the first quantization framework to enable both dequantization-free static mixed-precision of the backbone network, and truly efficient dynamic, instance-based quantization through a lightweight controller that decides at runtime how to quantize each layer. We demonstrate DQT state-of-the-art performance on ResNet18 on CIFAR-10 and ResNet50 on ImageNet. On ImageNet, our 4-bit dynamic ResNet50 achieves 77.00\% top-1 accuracy, an improvement over leading static (LSQ, 76.70\%) and dynamic (DQNET, 76.94\%) methods at a comparable BitOPs budget. Crucially, DQT achieves this with a bit-width transition cost of only 28.3M simple bit-shift operations, a drastic improvement over the 56.6M costly Multiply-Accumulate (MAC) floating-point operations required by previous dynamic approaches - unlocking a new frontier in efficient, adaptive AI.
\end{abstract}


\glsresetall

\section{Introduction}

The deployment of deep neural networks (DNNs) on resource-constrained hardware is a standard practice, enabled by quantization techniques that represent weights and activations with low-precision integers~\cite{nagel2021white, jacob2018quantization}. The dominant paradigm is static quantization, where a single, fixed bit-width (e.g., INT8) is used for inference on all inputs. This approach is suboptimal in resource-constrained settings as it allocates a fixed computational budget to every input, regardless of its intrinsic complexity.

Dynamic, instance-aware quantization addresses this limitation by adapting the network precision at runtime based on input characteristics~\cite{liu2022instance}. This method can achieve a superior accuracy-efficiency trade-off by allocating greater computational resources only to more challenging inputs. However, a critical and unresolved performance bottleneck has prevented its adoption. Existing dynamic methods depend on floating-point arithmetic to change precision, requiring a computationally expensive cycle of dequantization to FP32 followed by requantization to a target integer format INT(b) for each layer and input~\cite{xu2018dnq}. This reliance on floating-point operations breaks the efficient integer-only execution model of modern accelerators and introduces an overhead that undermines the intended efficiency gains.

This paper introduces a general framework to eliminate this dequantization bottleneck. Our core contribution is a nested integer representation where a lower-precision integer is a bit-wise truncation of its higher-precision counterpart. This property, combined with a co-designed set of integer-only arithmetic operators, enables bit-width transitions from a high precision $b_1$ to a lower precision $b_2$ via a single logical right bit-shift operation: $q_{b_2} \equiv q_{b_1} \gg (b_1-b_2)$. This mechanism replaces the expensive floating-point conversion cycle entirely. While this dequantization-free approach provides efficiency benefits for static mixed-precision and standard quantized networks, its impact is most significant in the dynamic, instance-aware setting.

We apply this general framework to create Dynamic Quantization Training (DQT), a complete end-to-end pipeline for training and deploying truly efficient adaptive models. In the DQT architecture, a lightweight controller adaptively selects per-layer bit-widths for each input. At inference, the model loads high-precision weights once and uses bit-shifting to instantiate the desired precision on-the-fly. The entire forward pass, including all precision adjustments, is executed using our custom integer arithmetic, maintaining a pure integer dataflow.

We validate DQT on the CIFAR-10 and ImageNet classification benchmarks. Our models establish a new state-of-the-art (SotA) regarding the accuracy-efficiency trade-off. For instance, our DQT-trained ResNet-50 operates at a computational budget lower than a fixed 4-bit network while achieving 77.00\% top-1 accuracy on ImageNet. By solving the dequantization bottleneck, DQT makes instance-aware quantization a practical and powerful tool for efficient AI. Our main contributions are:
\begin{enumerate}
    \item A general quantization framework based on a nested integer representation and custom integer-only arithmetic. This framework eliminates the floating-point conversion bottleneck common to all mixed-precision systems and enables bit-width changes via low-cost logical shift operations.
    \item The first application of this framework to develop an end-to-end training and inference pipeline, DQT, for dynamic, instance-aware networks. We demonstrate that by removing the dequantization overhead, DQT makes adaptive inference computationally practical and achieves state-of-the-art results.
\end{enumerate}

The rest of the paper is organized as follows. We first review prior work and provide background on uniform quantization. We then introduce our DQT framework and detail its core technical contribution. Subsequently, we describe the training and inference procedure, followed by extensive experimental results on CIFAR-10 and ImageNet. We conclude with a summary of our findings and directions for future work.
\section{Related Literature}
\label{sec:related_literature}

Our work is positioned at the intersection of static quantization, dynamic instance-aware quantization, and efficient integer arithmetic.

\paragraph{Static Quantization.}
The most common approach to model compression is static, uniform quantization, where weights and activations are converted to a fixed low bit-width (e.g., INT8) either after training (Post-Training Quantization, PTQ) or during it (Quantization-Aware Training, QAT)~\cite{nagel2021white, jacob2018quantization}. QAT is the standard for high-performance models, as it allows the network to compensate for quantization error. To improve upon uniform precision, static mixed-precision quantization assigns different bit-widths to different layers based on their sensitivity~\cite{dong2019hawqhessianawarequantization, wang2019haqhardwareawareautomatedquantization}. While these methods offer a better accuracy-efficiency trade-off, they remain fundamentally static; the computational cost is fixed for all inputs.

\paragraph{Dynamic, Instance-Aware Quantization.}
To address the rigidity of static methods, dynamic instance-aware quantization aims to adapt the computational cost to each input sample. Approaches in this area typically use a lightweight controller or policy network to select bit-widths at runtime. For instance, early work used reinforcement learning to select layer precisions~\cite{xu2018dnq}, while more recent methods like DQNet~\cite{liu2022instance} and AdaBit~\cite{jin2020adabits} employ small controllers trained jointly with the main network. These methods successfully demonstrate that adapting precision to input difficulty is feasible.

However, this entire class of methods is constrained by a critical performance issue: the dequantization bottleneck. To adjust a layer precision, these frameworks must dequantize the integer tensor to a 32-bit floating-point representation and subsequently requantize it to the new target bit-width. This dequantize-requantize cycle, performed per-layer and per-input, introduces substantial computational and memory-access overhead. This reliance on floating-point arithmetic breaks the optimized integer-only dataflow of modern accelerators, and the resulting overhead undermines the efficiency gains that dynamic precision is intended to provide. Consequently, instance-aware quantization has remained a largely theoretical exercise rather than a practical, deployable solution.

\paragraph{Nested Quantization and Integer Arithmetic.}
Recognizing this efficiency challenge, some research has explored alternative quantization schemes. Matryoshka Quantization (MatQuant)~\cite{nair2025matryoshka} introduced a nested quantization scheme where low-bit representations are embedded within high-bit ones. 
However, the goal of MatQuant is to extract multiple, independent static models of varying precisions from a single trained model. It relies on standard \textit{fake quantization} during training and does not provide the integer-only arithmetic necessary to perform dynamic, on-the-fly precision adjustments within a single inference pass.

Our work is the first to bridge this gap. We extend the nested representation concept by introducing a co-designed set of integer-only arithmetic operators compatible with bit-shift-based precision changes. This combination eliminates the dequantization bottleneck entirely, finally making efficient, instance-aware dynamic inference practical.

\section{Background: The Dequantization Bottleneck}
\label{sec:background}

To formalize the problem our framework solves, we first review the principles of standard uniform quantization and then detail the computational bottleneck that arises when applying it in a dynamic, mixed-precision context.

Uniform quantization maps a real-valued tensor \textit{x} from a clipping range $[m,M]$ to a \textit{b}-bit integer tensor using the quantization function $Q^b(\cdot)$:
\begin{equation}
    Q^b(x) = \text{clip}\left(\left\lfloor\frac{x-m}{\Delta}\right\rceil, 0, 2^b-1\right), \, \Delta = \frac{M-m}{2^b - 1}
    \label{eq:quant}
\end{equation}
with $\Delta$ being the quantization step.
The corresponding dequantization $D(\cdot)$ function reconstructs the real value $\hat{x}$:
\begin{equation}
    \hat{x}= D(x) = Q^b(x)\cdot \Delta + m
\end{equation}
The quantization error, $|x - \hat{x}|$, is bounded by the step size $\Delta$ and is the primary source of accuracy degradation in quantized models.

The rounding and clipping operations in \Cref{eq:quant} are non-differentiable, which prevents gradient-based optimization. Quantization-Aware Training (QAT) circumvents this issue using the Straight-Through Estimator (STE)~\cite{bengio2013estimating}. In the forward pass, a \textit{fake quantization} step is performed where a tensor is quantized and immediately dequantized ($x \to Q^b(x) \to \hat{x}$), simulating the precision loss. In the backward pass, the STE approximates the gradient of the quantization function as an identity, i.e., $\frac{\partial \hat{x}}{\partial x} = 1$. This allows gradients to flow through the quantization node to update the underlying full-precision weights.

\begin{figure*}[t]
    \centering
    \includegraphics[width=\linewidth]{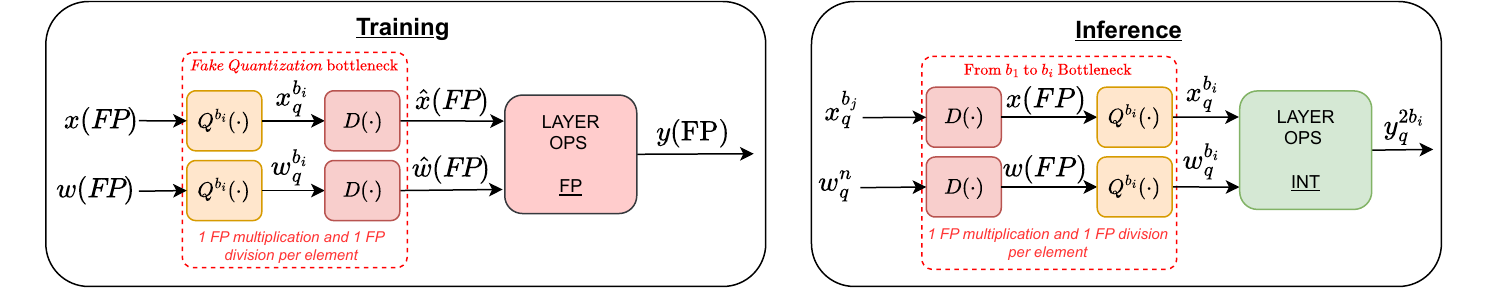}
    \caption{
    The dequantization bottleneck in conventional quantization frameworks at training time (left) and inference time (right). (Left) In standard QAT, dynamic precision requires a simulated dequantize-requantize cycle. (Right) At inference, transitioning between different integer bit-widths also necessitates this costly conversion to a floating-point intermediate. Our DQT framework eliminates this bottleneck in both scenarios.
    }
    \label{fig:qat-bottlenecks}
    \hspace{1pt}
    \includegraphics[width=\linewidth]{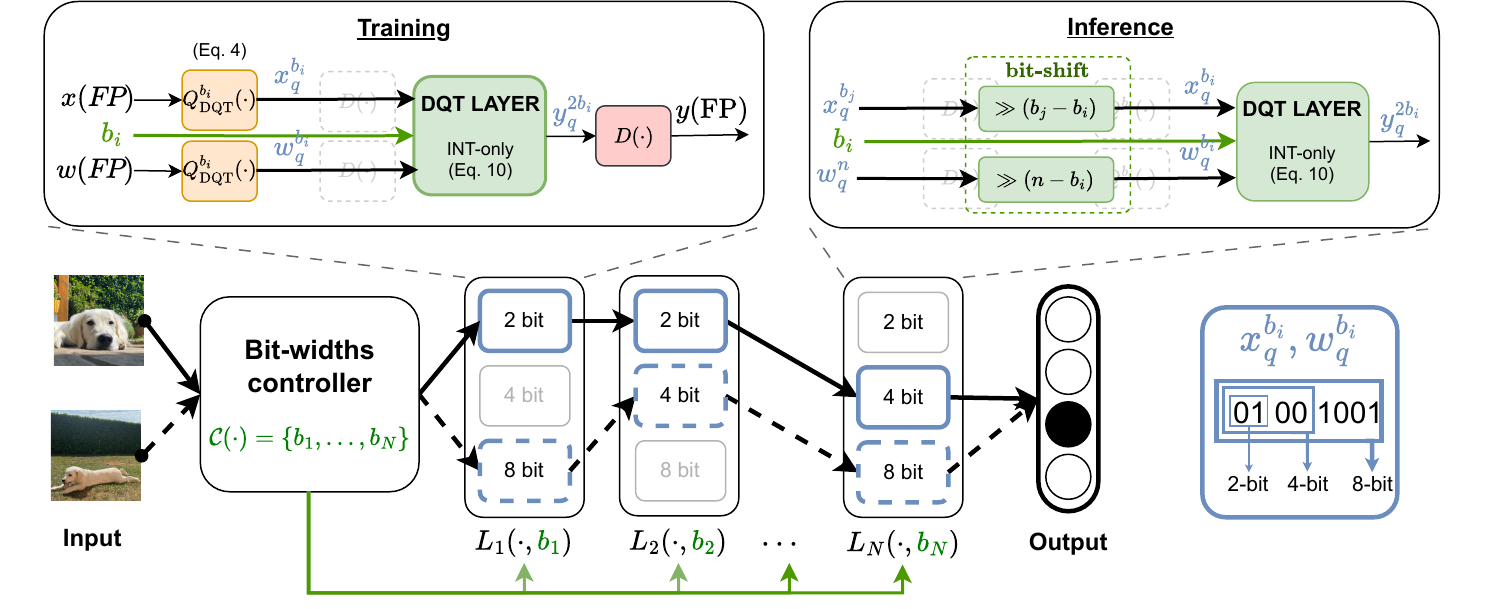}
    \caption{
    Overview of the Dynamic Quantization Training (DQT) framework. For each input, a lightweight controller $\mathcal{C}$ predicts per-layer bit-widths $\{b_i\}$. The backbone network operates on integer tensors, generating the required precision for weights ($W_q$) and activations ($x_q$) on-the-fly from a master representation via bit-shifting. The layer operations $L_i$ are executed using the dequantization-free integer arithmetic defined in Section~\ref{sec:method}.
    }
    \label{fig:architecture-overview}
\end{figure*}

The standard QAT process is designed for a single, static bit-width \textit{b}. This paradigm is computationally inefficient for dynamic, instance-aware models that must switch between different bit-widths at runtime. Consider changing a tensor precision from $b_1$ to $b_2$, with corresponding quantization parameters $(\Delta_1, m_1)$ and $(\Delta_2, m_2)$. In the standard framework, this requires a full dequantize-requantize cycle, as illustrated in~\Cref{fig:qat-bottlenecks}.
First, the tensor must be dequantized to a 32-bit floating-point representation using its current parameters:
$
    \hat{x} = Q^{b_1}(x) \cdot \Delta_1 -m_1
$. 
Then, it must be requantized to the new target precision using the new parameters:
$
    Q^{b_2}(\hat{x}) = \text{clip}\left(\left\lfloor\frac{\hat{x}-m_2}{\Delta_2}\right\rceil, 0, 2^{b_2}-1\right)
$. 
This dequantize-requantize cycle is the fundamental bottleneck. It requires performing expensive floating-point operations for every layer whose precision is changed, for every input sample. This computational overhead breaks the integer-only dataflow on hardware accelerators and compromises the performance benefits of using lower-precision integers. Our work eliminates this bottleneck entirely.

\section{DQT Dequantization-Free Quantization Framework}
\label{sec:method}


Quantized integer arithmetic is inconsistent with quantized floating-point results, i.e., $Q(x_1) \bullet Q(x_2) \ne Q(x_1 \bullet x_2)$, requiring dequantization. We eliminate this bottleneck via a novel framework built on two core principles: a nested integer representation that enables precision changes via bit-shifting, and custom bit-shift-compatible integer operators compatible with this representation.

The central mechanism of our framework is a nested quantization scheme where lower-precision representations are bit-wise embedded within higher-precision ones. We define a master bit-width, $n$, and a corresponding master scale,~$\Delta_n$. The quantization scale for any other bit-width, $b < n$, is defined by a power-of-two relationship:
\begin{equation}
    \Delta_b = \Delta_n \cdot 2^{n-b}
    \label{eq:scale_relation}
\end{equation}
This specific constraint on the quantization scales allows for a direct, dequantization-free conversion from a high-precision integer $Q^n(x)$ to a lower-precision integer $Q^b(x)$ via a single logical right bit-shift:
\begin{equation}
    \label{eq:bitshift_conversion}
    \begin{aligned}
        Q^b_\text{DQT}(x) &= \text{clip}\left(\left\lfloor \frac{Q^n(x)}{2^{n-b}} \right\rceil, 0, 2^b - 1\right) \\
                          &= Q^n(x) \gg (n - b) + \epsilon
    \end{aligned}
\end{equation}
where $\epsilon$ denotes the approximation error of the bit-shift operation and is analyzed in \Cref{appendix:error_analysis}.
This operation is computationally inexpensive and avoids any floating-point arithmetic. To leverage this property throughout a network, all arithmetic operations must be reformulated to operate directly on these integer representations and their corresponding scales.

We now present the integer-only arithmetic for addition and multiplication that is compatible with our nested quantization scheme. Let two real values, $x_1$ and $x_2$, be quantized to integers $q_1$ and $q_2$ with parameters $(\Delta_1, m_1)$ and $(\Delta_2, m_2)$ respectively, where $\Delta$ and $m$ follow the definitions from Section \ref{sec:background}. The output of an operation, $y = x_1 \bullet x_2$, will be quantized with parameters $(\Delta_y, m_y)$. 
The dequantized value of an input $q$ is $\hat{x} = q\cdot\Delta + m$. 
The goal is to compute the quantized output $q_y = \lfloor \frac{y-m_y}{\Delta_y} \rceil$ using only integer operations.

\paragraph{Integer Addition ($\oplus$):}

For addition, $y = x_1 + x_2$, the exact quantized output can be reformulated into a pure integer form:
\begin{equation}
    q_1 \oplus q_2 = k_1 q_1 + k_2 q_2 + k_3
    \label{eq:integer_add}
\end{equation}
where the scaling factors $k_i$ are pre-computed numbers:
\begin{equation}
    k_1 =  \left\lfloor\frac{\Delta_1}{\Delta_y}\right\rceil, \quad k_2 =  \left\lfloor \frac{\Delta_2}{\Delta_y}\right\rceil, \quad k_3 =  \left\lfloor\frac{m_1 +m_2-m_y}{\Delta_y}\right\rceil
    \label{eq:constants_add}
\end{equation}
The round operation $\left\lfloor\cdot\right\rceil$ on $k_i$ allows keeping the entire computation of the quantized output in the integer domain.

\paragraph{Integer Multiplication ($\otimes$):}
For multiplication, $y = x_1 \times x_2$, the exact quantized output leads to the integer-only formulation:
\begin{equation}
    q_1 \otimes q_2 = k_1q_1q_2 + k_2q_1 + k_3q_2 +k_4
    \label{eq:integer_mult}
\end{equation}
where the scaling factors $k_i$ are pre-computed numbers:
\begin{equation}
\begin{aligned}
    k_1 &= \left\lfloor\frac{\Delta_1 \Delta_2}{\Delta_y}\right\rceil, \quad
    k_2 = \left\lfloor \frac{\Delta_1 m_2}{\Delta_y} \right\rceil, \\
    k_3 &= \left\lfloor \frac{\Delta_2 m_1}{\Delta_y} \right\rceil, \quad
    k_4 = \left\lfloor \frac{m_1 m_2 - m_y}{\Delta_y} \right\rceil
\end{aligned}
\label{eq:constants_mult}
\end{equation}
The derivation of Eqs.~\eqref{eq:integer_add} and~\eqref{eq:integer_mult} together with the constants in Eqs.~\eqref{eq:constants_add} and~\eqref{eq:constants_mult} are reported in Appendix~\ref{appendix:A}.

\paragraph{Dynamic-Aware Scale Management.} The key to enabling dynamic precision is that the multipliers $k$ are simple ratios of scales. Due to the power-of-two relationship in \Cref{eq:scale_relation}, changing the bit-width of any input or output from~$b_1$ to~$b_2$ corresponds to multiplying or dividing its scale~$\Delta$ by~$2^{|b_1-b_2|}$. This means the multipliers~$k$ can also be updated with a simple, low-cost bit-shift at runtime, rather than being re-computed with floating-point division. 

\subsection{Integer-Only Layer Operations and Scale Management}
\label{sec:layer_ops}

We now apply the integer arithmetic from Section~\ref{sec:method} to construct complete neural network layers that operate in a dequantization-free manner. 
To execute the sequence of multiply-accumulate (MAC) operations required by NN layers in the integer domain, we must manage the quantization parameters $(\Delta, m)$ for weights, activations, and layer outputs in a way that is compatible with our nested quantization scheme.

\paragraph{Layer Formulation.} A floating-point layer $L_\mathrm{fp}(x, W, c)$ is reformulated into its integer-only equivalent, $L_\mathrm{int}(x_q, W_q, c_q)$. The core MAC operations are performed using the integer multiplication~($\otimes$) and addition~($\oplus$) operators defined in \Cref{eq:integer_add,eq:integer_mult}. The inputs $x_q$, weights $W_q$, and bias $c_q$ are integer tensors quantized to a dynamically selected bit-width $b$. The final integer output $y_q$ approximates the correctly quantized result of the original floating-point layer:
\begin{equation}
    \label{eq:int_layer}
    y_q = L_\mathrm{INT}(x_q, W_q, c_q) =  Q^b\bigl(L_\mathrm{FP}(x, W, c)\bigr) + \epsilon
\end{equation}
where $\epsilon$ denotes the approximation error of our formulation and is analyzed in \Cref{appendix:error_analysis}.

\paragraph{Weight and Bias Quantization.}
The weights $W$ and biases $c$ are static after training. We determine their clipping range $[m,M]$ and compute their master quantization steps $\Delta_{W,n}$ and $\Delta_{c,n}$ for the master bit-width $n$. These parameters are stored. At runtime, if a bit-width $b < n$ is selected, the required scale is derived directly using the power-of-two relationship from \Cref{eq:scale_relation}: $\Delta_{W,b} = \Delta_{W,n} \cdot 2^{n-b}$. 

\paragraph{Activation Quantization.}
Activation ranges are dynamic and input-dependent. To establish a stable quantization range, we adopt a PACT-like approach~\cite{choi2018pact}, where activations are clipped to a range $[0, \alpha]$. The clipping bound $\alpha$ is a learnable parameter optimized via gradient descent during training. This provides a fixed range from which we compute a stable master scale for activations, $\Delta_{A,n}$, ensuring compatibility with our nested scheme.

\paragraph{Layer Output Quantization.}
To determine the quantization parameters for a layer output, $(\Delta_{Y}, m_{Y})$, we must estimate its output range. This is done dynamically during training by tracking the running minimum and maximum values of the layer output using an Exponential Moving Average (EMA). For a layer output tensor~$y^{(t)}$ at training step~$t$, we update:
$
    y_\mathrm{max}^{(t)} = \gamma y_\mathrm{max}^{(t-1)} + (1 - \gamma) \max(y^{(t)})
$ and $
    y_\mathrm{min}^{(t)} = \gamma y_\mathrm{min}^{(t-1)} + (1 - \gamma) \min(y^{(t)})
$, 
where $\gamma$ is the EMA momentum. After training, the final range $[y_\mathrm{min}, y_\mathrm{max}]$ is used to compute the master output scale $\Delta_{Y,n}$.

This hierarchical scale management ensures that all quantization scales required for a layer computation - for weights, activations, and outputs - are defined relative to the master bit-width $n$. Consequently, all fixed-point multipliers $k$ needed for the integer arithmetic can be adjusted for any target bit-width $b$ with efficient bit-shifts, enabling a truly dynamic, integer-only layer execution. 

\section{The Dynamic Quantization Training (DQT) Framework}
\label{sec:dqt_framework}

This section presents Dynamic Quantization Training (DQT), an end-to-end framework for efficient, instance-aware inference. DQT leverages the dequantization-free methods from Section~\ref{sec:method} and integrates them into a complete system architecture. The system comprises two main components: a \textit{dequantization-free backbone network} that performs all computations in the integer domain, and a lightweight \textit{dynamic controller} that predicts per-layer bit-widths for each input instance. An overview of the DQT architecture is shown in \Cref{fig:architecture-overview}.

\subsection{Dequantization-Free Backbone Network}
\label{sec:backbone}

The backbone network executes the primary DNN computation. Each layer in the backbone, including \textit{convolution}, \textit{fully-connected (FC)}, \textit{batch normalization}, \textit{ReLU}, and \textit{skip connections}, has been completely re-implemented to operate in the nested integer domain. This is achieved by building each layer upon the DQT integer-only operations and scale management techniques detailed in Section~\ref{sec:layer_ops}.
This design allows each layer $L_i$ to operate on integer inputs and weights and to dynamically adapt to a bit-width $b_i$ selected by the controller.

The core of this adaptability is the nested quantization scheme. All weights are stored at a single master bit-width,~$n$, and all activations are processed relative to this precision. At runtime, a lower target precision $b_i$ for any tensor $x$ is generated on-the-fly via the bit-shift operation from \Cref{eq:bitshift_conversion}:
$Q_\text{DQT}^{b_i}(x) = Q^{n}(x) \gg (n - b_i)$.
This approach is both memory- and compute-efficient: only the $n$-bit weights require storage, and bit-width transitions require a single low-cost logical operation, completely eliminating the dequantize-requantize overhead.

The forward pass of each layer $L_i$ is therefore an integer-only function $L_{\mathrm{int}, i}$ that operates on tensors dynamically quantized to the chosen bit-width $b_i$:
\begin{equation}
    \label{eq:forward_dqt}
    y_q^{(i)} = L_{\mathrm{int}, i} \left( Q_\text{DQT}^{b_i}(x_q^{(i-1)}), Q_\text{DQT}^{b_i}(W_i), Q_\text{DQT}^{b_i}(c_i) \right)
\end{equation}
This design ensures that the entire forward pass, including all dynamic precision adjustments, maintains a purely integer dataflow.

\begin{algorithm}[t!]
    \caption{DQT End-to-End Training Procedure}
    \label{alg:training}
    \begin{algorithmic}[1]
    \Input Training data $(\mathcal{D})$, backbone with FP32 weights $W$, controller $\mathcal{C}$ with weights $W_\mathcal{C}$, hyperparameters $\alpha, \beta$, master bit-width $n$, candidate bit-widths $\{d_k\}$.
    \Output Trained weights $W$ and $W_\mathcal{C}$.
    
    \State Initialize $W$, $W_\mathcal{C}$.
    \For{each training epoch}
        \For{each batch $(X, Y) \in \mathcal{D}$}
            \State $p \leftarrow \mathcal{C}(X); \quad b_\text{dyn} \leftarrow \text{GumbelSoftmax}(p)$ 
            \State $Y_\text{pred} \leftarrow \text{ForwardPass}(X, W, b_\text{dyn}, n, \text{is\_train=True})$
            \State $J \leftarrow J_\mathrm{task} + \alpha J_\mathrm{con.cy} + \beta J_\mathrm{cost}$ \Comment{\Cref{eq:loss}} 
            \State Update $W$ and $W_\mathcal{C}$ with gradients $\nabla J$.
        \EndFor
    \EndFor
    \end{algorithmic}
\end{algorithm}

\begin{algorithm}[t!]
    \caption{DQT Dequantization-Free Inference}
    \label{alg:inference}
    \begin{algorithmic}[1]
    \Function{Inference}{$X_\text{sample}, W_q^n, \mathcal{C}, n$}
        \State $p \leftarrow \mathcal{C}(X_\text{sample}); \quad b \leftarrow \text{argmax}(p)$ 
        \State $Y_\text{pred} \leftarrow \text{ForwardPass}(X_\text{sample}, W_q^n, b, n, \text{is\_train=False})$
        \State \Return $Y_\text{pred}$
    \EndFunction
    \Statex
    \Function{ForwardPass}{$X, W, b, n, \text{is\_train}$}
        \State $W_q \leftarrow Q^n(W)$ if is\_train, else $W_q \leftarrow W$
        \State $x_q^{(0)} \leftarrow Q^n(X)$ 
        \For{$i = 1, \dots, N$} 
            \State $W_{q, b_i}^{(i)} \leftarrow W_q^{(i)} \gg (n - b_i)$ 
            \State $x_{q, b_i}^{(i-1)} \leftarrow x_q^{(i-1)} \gg (n - b_i)$ 
            \State $x_q^{(i)} \leftarrow L_{\mathrm{int}, i}(x_{q, b_i}^{(i-1)}, W_{q, b_i}^{(i)})$ \Comment{\Cref{eq:forward_dqt}}
        \EndFor
        \State \Return Dequantize$(x_q^{(N)})$ 
    \EndFunction
    \end{algorithmic}
\end{algorithm}
\subsection{Dynamic Bit-Width Controller}
\label{sec:controller}

The dynamic controller~$\mathcal{C}$ is a lightweight neural network that selects per-layer bit-widths for each input $\mathbf{x}$ for each of the $N$ backbone layers. It outputs logits for each of these layers, producing a probability distribution $\mathbf{p}_i = \text{softmax}(\mathcal{C}(\mathbf{x})_i)$ over $K$ candidate bit-widths ${d_1, \dots, d_K}$, where $\mathbf{p}_i \in \mathbb{R}^K$ is the probability vector for layer $L_i$.. 
At inference, bit-widths $b_i$ are chosen via a non-differentiable $\text{argmax}$ operation on the output logits. To enable end-to-end gradient-based training, we employ the Gumbel-Softmax reparameterization~\cite{jang2017categorical}, providing a differentiable approximation of the discrete selection process, allowing for joint optimization of the controller~$\mathcal{C}$ and the backbone network using standard backpropagation.

\begin{table*}[t!]
    \centering
    \begin{tabular}{llccccc}
    \toprule
    \textbf{Method} & \textbf{Model} & \textbf{W-Bits} & \textbf{A-Bits} & \textbf{BitOPs(G)} &\textbf{BW Transition Cost}\textsuperscript{*} & \textbf{Top-1 (\%)} \\
    \midrule
    \multicolumn{7}{c}{\textbf{--- CIFAR-10 Dataset ---}} \\
    \midrule
    DoReFa \cite{zhou2016dorefa} & ResNet18 & 3 & 3 & 0.34 &$\varnothing$& 80.90 \\
    PACT  \cite{choi2018pact} & ResNet18 & 3 & 3 & 0.34 &$\varnothing$& 91.10 \\
    DQNET \cite{liu2022instance}\textsuperscript{\textdagger} & ResNet18 & $\sim$3 MP & $\sim$3 MP & 0.36 &26.4M FLOPs& \underline{91.38} \\
    \textbf{Ours}\textsuperscript{\textdagger} & ResNet18 & $\sim$3 MP & $\sim$3 MP & 0.36 & 13.2M bit-shifts& \textbf{91.65} \\
    \midrule
    DoReFa \cite{zhou2016dorefa} & ResNet18 & 4 & 4 & 0.61 &$\varnothing$& 90.50 \\
    PACT  \cite{choi2018pact} & ResNet18 & 4 & 4 & 0.61 &$\varnothing$& 91.30 \\
    DQNET \cite{liu2022instance}\textsuperscript{\textdagger} & ResNet18 & $\sim$4 MP & $\sim$4 MP & 0.65 &26.4M FLOPs& \underline{91.60} \\
    \textbf{Ours}\textsuperscript{\textdagger} & ResNet18 & $\sim$4 MP & $\sim$4 MP & 0.64 & 13.2M bit-shifts& \textbf{91.73} \\
    \midrule
    DoReFa \cite{zhou2016dorefa} & ResNet18 & 5 & 5 & 0.95 &$\varnothing$& 90.40 \\
    PACT  \cite{choi2018pact} & ResNet18 & 5 & 5 & 0.95 & $\varnothing$&91.70 \\
    DQNET \cite{liu2022instance}\textsuperscript{\textdagger} & ResNet18 & $\sim$5 MP & $\sim$5 MP & 0.98 &26.4M FLOPs& \underline{92.01} \\
    \textbf{Ours}\textsuperscript{\textdagger} & ResNet18 & $\sim$5 MP & $\sim$5 MP &  0.97 & 13.2M bit-shifts& \textbf{92.76} \\
    \midrule
    \multicolumn{7}{c}{\textbf{--- ImageNet Dataset ---}} \\
    \midrule
    DoReFa \cite{zhou2016dorefa} & ResNet50 & 4 & 4 & 61.27 &$\varnothing$& 71.40 \\
    PACT  \cite{choi2018pact} & ResNet50 & 4 & 4 & 61.27 &$\varnothing$&76.50 \\
    LQ-Nets \cite{zhang2018lqnets} & ResNet50 & 4 & 4 & 61.27 &$\varnothing$& 74.89 \\
    LSQ \cite{esser2020learnedstepsizequantization} & ResNet50 & 4 & 4 & 61.27 &$\varnothing$& 76.70 \\
    HAQ \cite{wang2019haqhardwareawareautomatedquantization} & ResNet50 & $\sim$4 MP & $\sim$4 MP & -&56.6M FLOPs& 76.14 \\
    DQNET \cite{liu2022instance}\textsuperscript{\textdagger} & ResNet50 & $\sim$4 MP & $\sim$4 MP & 61.49 &56.6M FLOPs& \underline{76.94} \\
    \textbf{Ours}\textsuperscript{\textdagger} & ResNet50 & $\sim$4 MP & $\sim$4 MP & 61.39  & 28.3M bit-shifts& \textbf{77.00} \\
    \midrule
    DoReFa \cite{zhou2016dorefa} & ResNet50 & 5 & 5 & 95.73 &$\varnothing$& 71.40 \\
    PACT  \cite{choi2018pact} & ResNet50 & 5 & 5 & 95.73 &$\varnothing$& 76.70 \\
    DQNET \cite{liu2022instance}\textsuperscript{\textdagger} & ResNet50 & $\sim$5 MP & $\sim$5 MP & 96.27 &56.6M FLOPs& \underline{77.12} \\
    \textbf{Ours}\textsuperscript{\textdagger} & ResNet50 & $\sim$5 MP & $\sim$5 MP & 95.97 & 28.3M bit-shifts& \textbf{77.25} \\
    \midrule
    MPDNN \cite{9712485} & MobileNetV2 & $\sim$4 MP & $\sim$4 MP & - & $\varnothing$ & 69.74 \\
    AutoQ \cite{lou2020autoqautomatedkernelwiseneural} & MobileNetV2 & $\sim$4 MP & $\sim$4 MP & -&$\varnothing$ & 70.80 \\
    FracBits \cite{yang2020fracbitsmixedprecisionquantization} & MobileNetV2 & $\sim$4 MP & $\sim$4 MP & -&$\varnothing$ & 71.30 \\
    DQNET \cite{liu2022instance}\textsuperscript{\textdagger} & MobileNetV2 & $\sim$4 MP & $\sim$4 MP & 2.81 &7.0M FLOPs& \textbf{72.05} \\
    \textbf{Ours}\textsuperscript{\textdagger} & MobileNetV2 & $\sim$4 MP & $\sim$4 MP & 2.65 &3.5M bit-shifts& \underline{72.01} \\
    \bottomrule
    \multicolumn{7}{l}{\textsuperscript{\textdagger} Instance-based architecture.} \\
    \multicolumn{7}{l}{\textsuperscript{*} Inference Bit-Width (BW) transition cost from the masted BW to the selected one in the worst-case.} \\
    \end{tabular}
    \caption{
    Comparison of Top-1 Accuracy (\%) and computational cost (BitOPs in G) on CIFAR-10 and ImageNet. DQT is compared against state-of-the-art static and dynamic quantization methods. For dynamic methods ($\dagger$), BitOPs are averaged over the validation set. BW Transition Cost is the operational overhead for changing bit-widths per inference. Our method replaces high-latency FLOPs with low-cost logical shifts. The best Top-1 accuracy is reported in \textbf{bold}, the second-best is \underline{underlined}.
    }
    \label{tab:results_main}
\end{table*}
\subsection{Training and Inference}
\label{sec:training_inference_dqt}

The DQT backbone and controller are trained jointly in an end-to-end fashion. The training objective is designed to optimize for task accuracy while regularizing for computational cost and model robustness.

\paragraph{Training and Inference Procedures.}
The complete end-to-end procedures for training and inference are outlined in Algorithm~\ref{alg:training} and Algorithm~\ref{alg:inference}, respectively. During training, the forward pass simulates the full dynamic process using the Gumbel-Softmax for bit-width selection. The backward pass computes gradients with respect to both $W$ and $W_\mathcal{C}$, using the Straight-Through Estimator (STE) for quantization nodes. At inference, the trained floating-point weights $W$ are quantized once to the master bit-width $n$ and stored. For each input, the controller deterministically selects the per-layer bit-widths, and the backbone performs a dequantization-free forward pass using bit-shifting and our custom integer arithmetic. Dequantization to floating-point occurs only once for the final network output.

\paragraph{Training Objective.}
We optimize a multi-term loss $J$ over the full-precision backbone weights $W$ and controller parameters $W_\mathcal{C}$:
$
    W^*, W_\mathcal{C}^* = \arg\min_{W, W_\mathcal{C}} J
$, 
where:
\begin{equation}
    \label{eq:loss}
    J = J_\mathrm{task} + \alpha J_\mathrm{consistency} + \beta J_\mathrm{cost}.
\end{equation}
$J_\mathrm{task}$ is the standard loss (e.g., cross-entropy) computed using the controller dynamic bit-widths.
$J_\mathrm{consistency}$ 
ensures the backbone network performs well across all candidate bit-widths, not just those frequently chosen by the controller. 
This term is the sum of $J_\mathrm{task}$ computed over a small, fixed set of uniform bit-width configurations, $K' \subset \{d_1, \dots, d_K\}$ (e.g., the minimum and maximum available bit-widths):
$
    J_\mathrm{consistency} = \sum_{b \in K'} J^{(b)}_\mathrm{task}
$, 
where $J^{(b)}_\mathrm{task}$ is the task loss when the entire network is statically set to bit-width~$b$. This stabilizes training by ensuring the shared weights are well-conditioned for any potential bit-width selection.
$J_\mathrm{cost}$ 
penalizes large the choice of large bit-widths by the controller, by computing the expected bit-width across all $N$ layers, averaged over the batch. 
$
    J_\mathrm{cost} = \frac{1}{N}\sum_{i=1}^{N} \sum_{k=1}^{K} p_{i,k} \cdot d_k
$, 
where $p_{i,k}$ is the Gumbel-Softmax probability of selecting bit-width $d_k$ for layer $i$.
The hyperparameters $\alpha$ and $\beta$ control the trade-off between accuracy and computational cost; their impact is analyzed in \Cref{appendix:alpha_beta_effect}.

\section{Experiments}
\label{sec:experiments}

We conduct a series of experiments to validate the performance of the DQT framework. Our primary hypothesis is that by eliminating the dequantization bottleneck, DQT can achieve a SotA accuracy-efficiency trade-off compared to both static and existing dynamic quantization methods.

\subsection{Experimental Setup}

We evaluate our method on two standard image classification benchmarks: CIFAR-10~\cite{Krizhevsky09learningmultiple} and the large-scale ImageNet ILSVRC 2012 dataset~\cite{ILSVRC15}. We use widely adopted architectures such as ResNet-18/50~\cite{he2015deepresiduallearningimage} and MobileNetV2~\cite{sandler2019mobilenetv2invertedresidualslinear}.
We compare DQT against a comprehensive set of baseline methods. For static uniform quantization, we include learned methods such as DoReFa-Net~\cite{zhou2016dorefa}, PACT~\cite{choi2018pact}, and LSQ~\cite{esser2020learnedstepsizequantization}. For static mixed-precision, we compare against HAWQ~\cite{dong2019hawqhessianawarequantization} and HAQ~\cite{wang2019haqhardwareawareautomatedquantization}. Our primary dynamic baseline is DQNet~\cite{liu2022instance}, the leading method for instance-aware dynamic quantization.

We implement all models in PyTorch and train them using SGD. Key hyperparameters, such as learning rates, schedules, weight decay, the controller structure, and the values for $\alpha$ and $\beta$, are detailed in Appendix~\ref{sec:hyperparameters} to ensure reproducibility.
We evaluate all models by Top-1 Accuracy (\%) on the test set and computational cost in Bit-Operations (BitOPs), computed as $\text{BitOPs} = \sum_i (\text{MACs}_i \cdot b_{w,i} \cdot b_{a,i})$, where~$b_{w,i}$ and~$b_{a,i}$ are the bit-widths for weights and activations of layer~$i$.
For dynamic models, BitOPs are averaged over the validation set. We also report the operational cost of bit-width transitions, a key point of comparison.

\subsection{Performance Analysis}
\label{sec:performance_analysis}

Table~\ref{tab:results_main} presents a comprehensive comparison of DQT against all baselines. Across all architectures and datasets, DQT consistently establishes a new SotA accuracy-efficiency trade-off.
On ImageNet with ResNet-50, our 4-bit dynamic DQT model achieves 77.00\% top-1 accuracy, outperforming the SotA static method LSQ (76.70\%) and the leading dynamic method DQNet (76.94\%) at a comparable BitOPs budget. This result demonstrates that DQT can exceed the performance of static quantization while retaining the benefits of adaptivity. Critically, the bit-width transitions in DQT require only 28.3M logical shift operations, whereas prior dynamic methods incur an overhead of 56.6M high-latency floating-point operations.
Similarly, on CIFAR-10 with ResNet-18, DQT consistently outperforms all baselines across 3, 4, and 5-bit configurations, achieving up to 92.76\% accuracy. 
This shift from floating-point arithmetic to logical operations is able to provide a substantial speedup across hardware, both on CPUs by reducing instruction latency and on GPUs by maintaining an efficient integer-only dataflow. We provide precise per-element cost estimations to quantify the computational advantage in \Cref{sec:dequantization_bottleneck_analysis}.


\paragraph{Memory Footprint.}
The DQT framework is designed for computational optimization. The model memory footprint is determined by the master bit-width $n$, as the full $n$-bit weights must be stored. We use $n=8$, that yields a $4\times$ memory reduction relative to FP32, which is standard for INT8 models. The storage overhead for the integer arithmetic multipliers and the controller parameters is negligible, constituting less than 1\% of the total model size. DQT therefore maintains a memory footprint comparable to a static high-precision model while enabling substantial savings in computational cost.

\paragraph{Computational Cost of Integer Arithmetic.}

While our framework eliminates floating-point operations, the custom integer operators in Section~\ref{sec:method} require more than a single hardware MAC. For instance, our integer multiplication (\Cref{eq:integer_mult}) involves three multiplications and an accumulation. 
To reduce this cost, we apply PACT~\cite{choi2018pact} clamping to the activations, setting the zero-point to zero and removing the $k_3$ term in Eq.~\ref{eq:integer_mult}, saving one multiplication.
Though our custom integer MAC requires 2–3 operations versus a single standard MAC, it fully avoids the dequantize-requantize cycle. 
This trade-off is advantageous for dynamic inference, where frequent bit-width switching would otherwise dominate the computational cost.

\section{Conclusion and Future Work}
\label{sec:conclusions}

This paper addressed a fundamental performance bottleneck that has hindered the practical application of dynamic, instance-aware quantization: the computational overhead of runtime dequantization and requantization. 
We introduced a dequantization-free quantization framework built on two core components: a nested integer representation and a compatible set of custom integer-only arithmetic operators. This design enables bit-width transitions via a single, low-cost logical bit-shift operation, eliminating the need for floating-point conversions. We applied this framework to develop Dynamic Quantization Training (DQT), an end-to-end pipeline for training and deploying efficient, adaptive~NNs.

Our experimental results on CIFAR-10 and ImageNet demonstrate that DQT establishes a new SotA considering the accuracy-efficiency trade-offs. By removing the dequantization bottleneck, DQT makes dynamic, instance-aware inference a computationally practical and superior alternative to static quantization. While our experiments focused on this most challenging application of dynamic quantization, the underlying dequantization-free arithmetic represents a general advance for any quantized system, offering a more efficient execution path for both static and dynamic mixed-precision models.

Future research can extend this work in several promising directions. 
A natural direction is to eliminate the remaining dependency in the training phase. While DQT enables integer-only inference, the backward pass still relies on floating-point gradients. A fully integer-based training approach, including quantized gradients and updates, would represent a complete solution for efficient deep learning on integer-only hardware.
The integer-only operators are well-suited for hardware co-design, with custom ASIC or FPGA instructions offering further speedups.
Extending this framework to other domains, including large language models (LLMs) and object detection, is another important direction. 
\bibliography{ref}

\setcounter{secnumdepth}{1}
\clearpage
\appendix

\appendix
\section{Derivation of Integer-Only Operators}
\label{appendix:A}

This appendix provides the mathematical derivation for the integer-only operators $\oplus$ (addition) and $\otimes$ (multiplication) and their corresponding constants, as introduced in Section~\ref{sec:method}. A real value $x$ is quantized to an integer~$q$ using a scale~$\Delta$ and the minimum of the quantization range,~$m$. The dequantized value is $\hat{x} = q \cdot \Delta + m$.

\subsection{Derivation of Integer Addition}

Let $x_1$ and $x_2$ be two real values, quantized to integers $q_1$ and $q_2$ with parameters $(\Delta_1, m_1)$ and $(\Delta_2, m_2)$ respectively. Their sum is $y = x_1 + x_2$. The goal is to compute the quantized representation of the sum, $q_y$, using only integer operations, where $q_y$ is defined with parameters $(\Delta_y, m_y)$.

From the definition of quantization in \Cref{eq:quant}, we have:
\begin{equation}
    q_y = \text{clip}\left(\left\lfloor\frac{y - m_y}{\Delta_y}\right\rceil, 0, 2^b-1\right)
\end{equation}
Substituting the dequantized values for $x_1$ and $x_2$, we get:
\begin{equation}
\begin{aligned}
    q_y &= \text{clip}\left(\left\lfloor \frac{(q_1\Delta_1+m_1) + (q_2\Delta_2+m_2) - m_y}{\Delta_y} \right\rceil, \dots \right)\\
    &= \text{clip}\left(\left\lfloor q_1\frac{\Delta_1}{\Delta_y} + q_2\frac{\Delta_2}{\Delta_y} + \frac{m_1 + m_2 - m_y}{\Delta_y} \right\rceil, \dots \right)
\end{aligned}
\end{equation}
To convert this expression into an integer-only form, we define integer constants $k_i$ by rounding the scaling ratios. This leads to the approximation:
\begin{equation}
    q_1 \oplus q_2 = \text{clip}(k_1 q_1 + k_2 q_2 + k_3, 0, 2^b-1)
\end{equation}
where the constants $k_i$ are pre-computed as defined in \Cref{eq:constants_add}:
\begin{equation}
    k_1 =  \left\lfloor\frac{\Delta_1}{\Delta_y}\right\rceil, \quad k_2 =  \left\lfloor \frac{\Delta_2}{\Delta_y}\right\rceil, \quad k_3 =  \left\lfloor\frac{m_1 +m_2-m_y}{\Delta_y}\right\rceil
\end{equation}
The rounding operation on the $k_i$ constants ensures the computation is performed entirely in the integer domain. The error introduced by this rounding is analyzed in Appendix~\ref{appendix:error_analysis}.

\subsection{Derivation of Integer Multiplication}

Similarly, for multiplication, the product is $y = x_1 \cdot x_2$. The goal is to compute its quantized representation $q_y$.
\begin{align}
 q_y &= \text{clip}\left(\left\lfloor \frac{(q_1\Delta_1+m_1) \cdot (q_2\Delta_2+m_2) - m_y}{\Delta_y} \right\rceil, \dots \right)\nonumber\\
  &= \text{clip}\bigg(\bigg\lfloor \frac{\Delta_1\Delta_2}{\Delta_y}q_1q_2 + \frac{\Delta_1m_2}{\Delta_y}q_1+\frac{\Delta_2m_1}{\Delta_y}q_2\nonumber \\
  & \quad \quad \quad \quad \quad + \frac{m_1m_2 - m_y}{\Delta_y} \bigg\rceil, \dots \bigg)
\end{align}
By defining integer constants $k_i$ for each term, we arrive at the integer-only approximation:
\begin{equation}
    q_1 \otimes q_2 = \text{clip}(k_1q_1q_2 + k_2q_1 + k_3q_2 +k_4, 0, 2^b-1)
\end{equation}
where the constants $k_i$ are pre-computed as defined in \Cref{eq:constants_mult}:
\begin{equation}
\begin{aligned}
    k_1 &= \left\lfloor\frac{\Delta_1 \Delta_2}{\Delta_y}\right\rceil, &
    k_2 &= \left\lfloor \frac{\Delta_1 m_2}{\Delta_y} \right\rceil \\
    k_3 &= \left\lfloor \frac{\Delta_2 m_1}{\Delta_y} \right\rceil, &
    k_4 &= \left\lfloor \frac{m_1 m_2 - m_y}{\Delta_y} \right\rceil
\end{aligned}
\end{equation}
This formulation allows the multiplication of two quantized numbers to be executed entirely with integer arithmetic. The final clipping operation ensures the result remains within the valid range for the target bit-width $b$.

\subsection{Derivation of Integer Dot Product}

The multiply-accumulate (MAC) operation, which forms the basis of dot products and convolutions, can be constructed from the integer operators $\oplus$ and $\otimes$. Here, we derive the integer-only formulation for the dot product between a quantized weight vector $\mathbf{w}_q$ and an activation vector $\mathbf{x}_q$.

Let the real-valued vectors be $\mathbf{x} = [x_1, \dots, x_N]$ and $\mathbf{w} = [w_1, \dots, w_N]$. Their quantized integer representations are $\mathbf{x}_q = [x_{q,1}, \dots, x_{q,N}]$ and $\mathbf{w}_q = [w_{q,1}, \dots, w_{q,N}]$, with quantization parameters $(\Delta_x, m_x)$ and $(\Delta_w, m_w)$, respectively. The dot product is $y = \sum_{i=1}^N x_i \cdot w_i$. We wish to compute its quantized representation $q_y$ with parameters $(\Delta_y, m_y)$.

The dequantized value of the dot product is:
\begin{equation}
\begin{aligned}
\hat{y} &= \sum_{i=1}^N (x_{q,i}\Delta_x + m_x)(w_{q,i}\Delta_w + m_w) \\
&= \sum_{i=1}^N \left[ x_{q,i}w_{q,i}\Delta_x\Delta_w + x_{q,i}\Delta_x m_w + w_{q,i}\Delta_w m_x + m_x m_w \right]
\end{aligned}
\end{equation}
Rearranging the terms, we get:
\begin{equation}
\begin{aligned}
\hat{y} = &(\Delta_x\Delta_w) \sum_{i=1}^N x_{q,i}w_{q,i} + (\Delta_x m_w) \sum_{i=1}^N x_{q,i} \\
& + (\Delta_w m_x) \sum_{i=1}^N w_{q,i} + N \cdot m_x m_w
\end{aligned}
\end{equation}
To quantize this result, we compute $q_y = \lfloor \frac{\hat{y} - m_y}{\Delta_y} \rceil$. Substituting $\hat{y}$ and rounding the scaling ratios to integer constants leads to the final integer-only dot product formulation:
\begin{equation}
\label{eq:integer_dot_product}
\begin{aligned}
q_{\text{dot}} = \text{clip} \Bigg( &k_1 \sum_{i=1}^N x_{q,i}w_{q,i} + k_2 \sum_{i=1}^N x_{q,i} \\
&+ k_3 \sum_{i=1}^N w_{q,i} + k_4, \dots \Bigg)
\end{aligned}
\end{equation}
where the pre-computed integer constants are:
\begin{equation}
\begin{aligned}
    k_1 &= \left\lfloor \frac{\Delta_x \Delta_w}{\Delta_y} \right\rceil, &
    k_2 &= \left\lfloor \frac{\Delta_x m_w}{\Delta_y} \right\rceil \\
    k_3 &= \left\lfloor \frac{\Delta_w m_x}{\Delta_y} \right\rceil, &
    k_4 &= \left\lfloor \frac{N \cdot m_x m_w - m_y}{\Delta_y} \right\rceil
\end{aligned}
\end{equation}
This formulation shows that a dot product can be implemented efficiently in the integer domain. It requires three main accumulation terms: $\sum x_{q,i}w_{q,i}$, $\sum x_{q,i}$, and $\sum w_{q,i}$. These terms are then scaled by the integer constants $k_i$ and summed. This structure makes the per-MAC operational cost explicit, consisting of one multiplication and several additions, which is the basis for the complexity analysis in our experimental section.

\paragraph{Implementation and Accumulator Management.}
The integer dot product formulation in \Cref{eq:integer_dot_product} has direct implications for hardware or software implementation. A primary consideration is managing the bit-width of the accumulators. Given $n$-bit integer inputs $x_{q,i}$ and $w_{q,i}$, their product $x_{q,i}w_{q,i}$ requires up to $2n$ bits. Accumulating $N$ such products for the term $\sum_{i=1}^N x_{q,i}w_{q,i}$ would require an accumulator with a bit-width of approximately $2n + \lceil\log_2(N)\rceil$ to prevent overflow.
This bit growth is a standard challenge in integer-based neural network accelerators. To manage this in our implementation, the accumulated $2n + \lceil\log_2(N)\rceil$-bit value is rescaled back to a smaller bit-width (e.g., 32-bit) before applying the final scaling factors~$k_i$. This rescaling is performed via a logical right bit-shift, effectively dividing the sum by a power of two. This is a common technique in digital signal processing and is analogous to the requantization step that follows a standard INT32 accumulation in conventional quantized inference.
\section{Bit-Width Transition via Logical Shift}
\label{appendix:bitshift}

This appendix provides a formal justification for the dequantization-free bit-width transition presented in Section~\ref{sec:method}. We show that by enforcing a power-of-two relationship between quantization scales, the conversion from a high bit-width integer to a low bit-width integer reduces to a simple logical right bit-shift.

Let a real value $x$ be quantized to an $n$-bit integer $q_n$ using the parameters $(\Delta_n, m_n)$. From \Cref{eq:quant}, we have:
\begin{equation}
    q_n = \left\lfloor\frac{x - m_n}{\Delta_n}\right\rceil
\end{equation}
We wish to find the corresponding $b$-bit integer, $q_b$, where $b < n$. The parameters for the $b$-bit representation are $(\Delta_b, m_b)$. For our nested quantization scheme, we enforce two constraints: (1) The quantization range is preserved, meaning the minimum value is the same: $m_b = m_n = m$; (2) The scales are related by a power-of-two, as defined in \Cref{eq:scale_relation}: $\Delta_b = \Delta_n \cdot 2^{n-b}$.

\begin{figure}[t!]
    \centering
    \includegraphics[width=\linewidth]{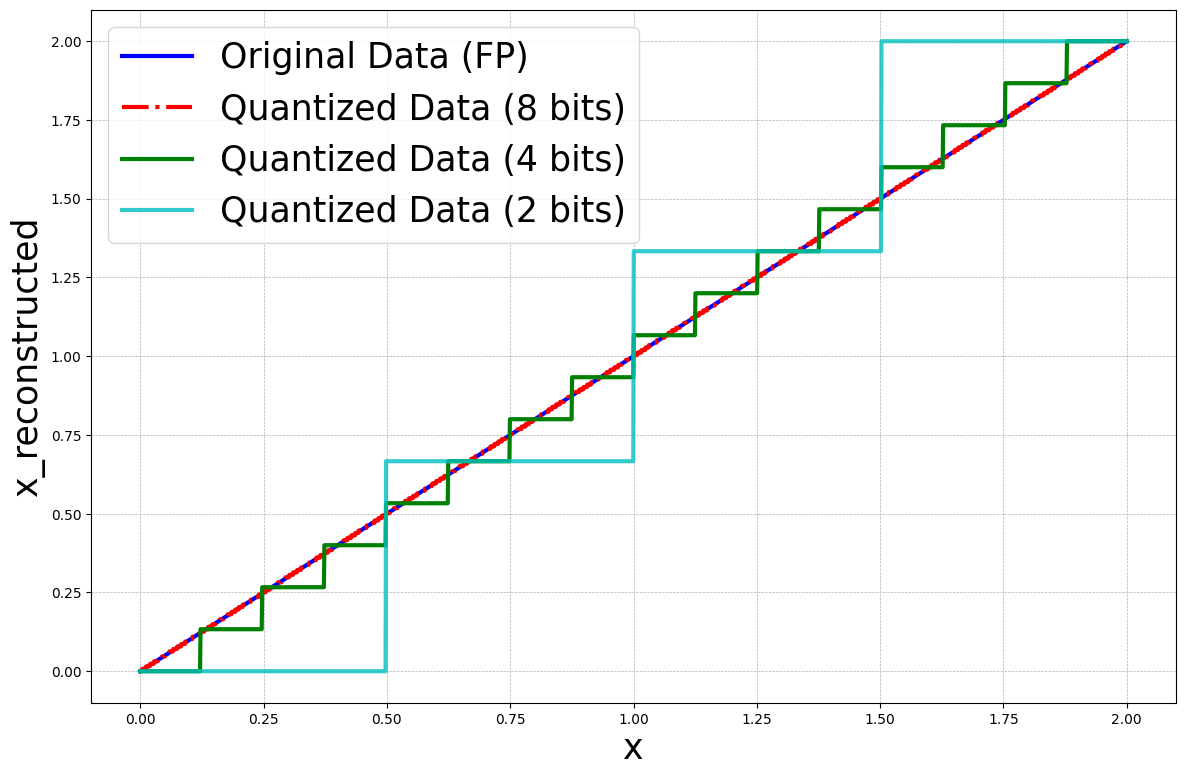}
    \caption{
    Visualization of the nested quantization scheme using the efficient bit-shifting approximation. The original floating-point (FP) data is first quantized to a master 8-bit representation ($q_8$). Lower-precision representations ($q_4$ and $q_2$) are then generated directly from the 8-bit integer via logical right bit-shifting~(\( \gg \)), demonstrating the embedding of low-precision values within the high-precision master.
    }
    \label{fig:quantization_comparison}
\end{figure}

Under these constraints, the expression for $q_b$ becomes:
\begin{equation}
    q_b = \left\lfloor\frac{x - m}{\Delta_b}\right\rceil = \left\lfloor\frac{x - m}{\Delta_n \cdot 2^{n-b}}\right\rceil
\end{equation}
From the definition of $q_n$, we know that the term $\frac{x - m}{\Delta_n}$ is approximately equal to the integer $q_n$. Substituting this relationship gives:
\begin{equation}
\begin{aligned}
    q_b &\approx \left\lfloor\frac{q_n}{2^{n-b}}\right\rceil
\end{aligned}
\end{equation}
This operation, dividing an integer by a power of two and rounding to the nearest integer, is equivalent to a logical right bit-shift for unsigned integers, followed by a final clipping to the valid $b$-bit range. This leads to our core bit-width transition mechanism, which is visualized in \Cref{fig:quantization_comparison}:
\begin{equation}
    Q^b_\text{DQT}(x) = \text{clip}\left( Q^n(x) \gg (n-b), 0, 2^b-1 \right)
\end{equation}
This derivation confirms that our specific scale constraint (\Cref{eq:scale_relation}) is precisely what enables the computationally efficient, dequantization-free transition between bit-widths. 
The error introduced by this integer-based approximation is analyzed in Appendix~\ref{appendix:error_analysis}.

\section{Approximation Error Analysis}
\label{appendix:error_analysis}

Our dequantization-free framework introduces two primary sources of approximation error, which are systematically compensated for during QAT.
This appendix provides a formal analysis of the distinct sources of approximation error that arise from the DQT design. 

\subsection{Error from Integer Operator Rounding}
\label{appendix:op_error}

The first source of error is introduced by the integer-only operators $\oplus$ and $\otimes$ defined in Appendix~\ref{appendix:A}. To maintain a pure integer dataflow, these operators rely on pre-computed integer constants $k_i$, which are obtained by rounding their exact real-valued counterparts. This rounding introduces a small, bounded error into each arithmetic operation.

Let the exact real-valued multiplier for a given term be $k_{i,\text{real}}$ and the corresponding rounded integer constant be $k_i = \lfloor k_{i,\text{real}} \rceil$. The rounding error for this single constant is defined as $\delta_i = k_{i,\text{real}} - k_i$, where $|\delta_i| \le 0.5$. The total approximation error for an operation is the difference between the exact quantized value (computed with $k_{i,\text{real}}$) and our integer-only approximation (computed with $k_i$). This error propagates through the operations as follows.

\begin{figure*}[h!]
    \centering
    \begin{subfigure}[b]{0.245\textwidth}
        \includegraphics[width=\textwidth]{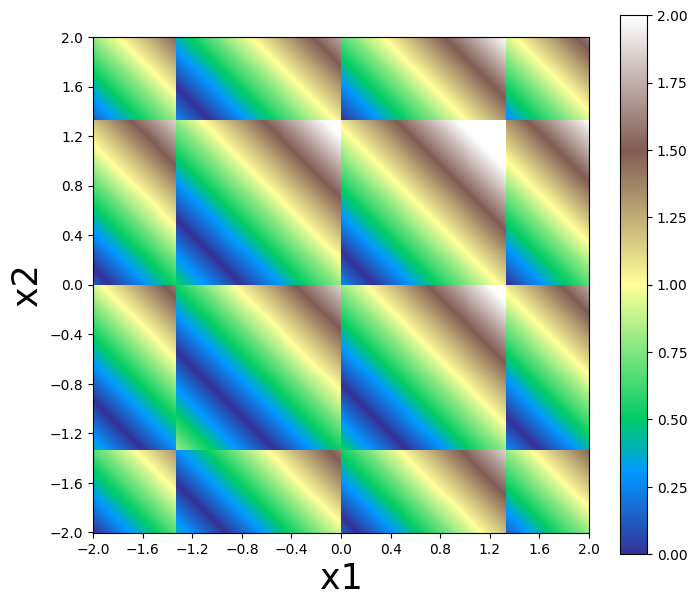}
        \caption{2 bit}
        \label{fig:sub1}
    \end{subfigure}
    \begin{subfigure}[b]{0.245\textwidth}
        \includegraphics[width=\textwidth]{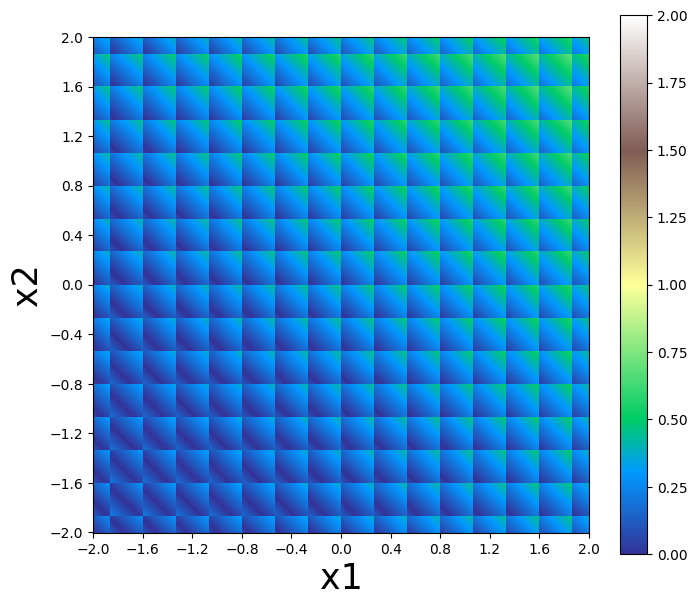}
        \caption{4 bit}
        \label{fig:sub2}
    \end{subfigure}
    \begin{subfigure}[b]{0.245\textwidth}
        \includegraphics[width=\textwidth]{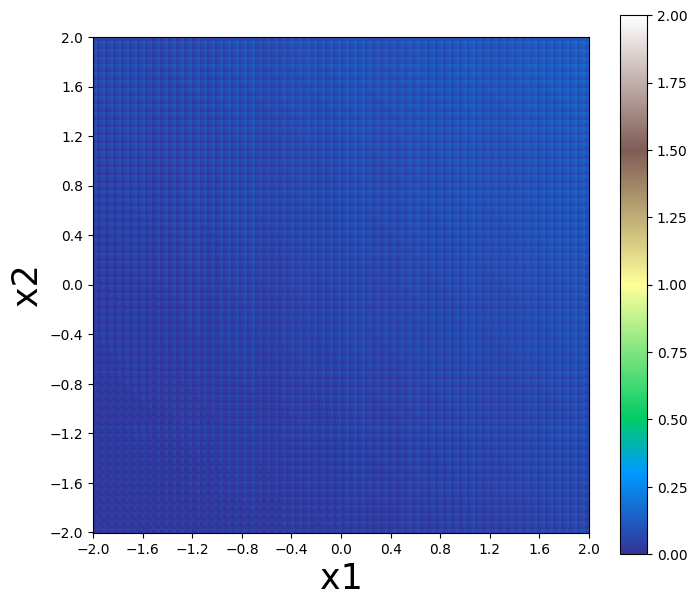}
        \caption{6 bit}
        \label{fig:sub3}
    \end{subfigure}
    \begin{subfigure}[b]{0.245\textwidth}
        \includegraphics[width=\textwidth]{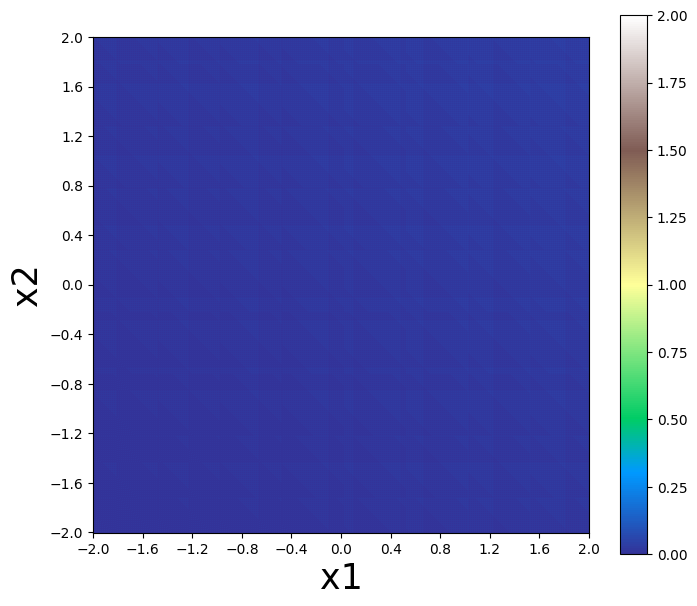}
        \caption{8 bit}
        \label{fig:sub4}
    \end{subfigure}
    \caption{Approximation error $\epsilon_\oplus$ (defined in \Cref{eq:error_sum}) of the DQT integer-only addition operator (see \Cref{appendix:A}) compared to standard floating-point addition $x_1 + x_2$, evaluated over the interval $[-2, 2]$ for varying bit-widths. The observed asymmetry around zero arises from the rounding of fixed constants in the DQT implementation; when rounding is removed, the error becomes symmetric.}
    \label{fig:sum_error}
\end{figure*}
\begin{figure*}[h!]
    \centering
    \begin{subfigure}[b]{0.245\textwidth}
        \includegraphics[width=\textwidth]{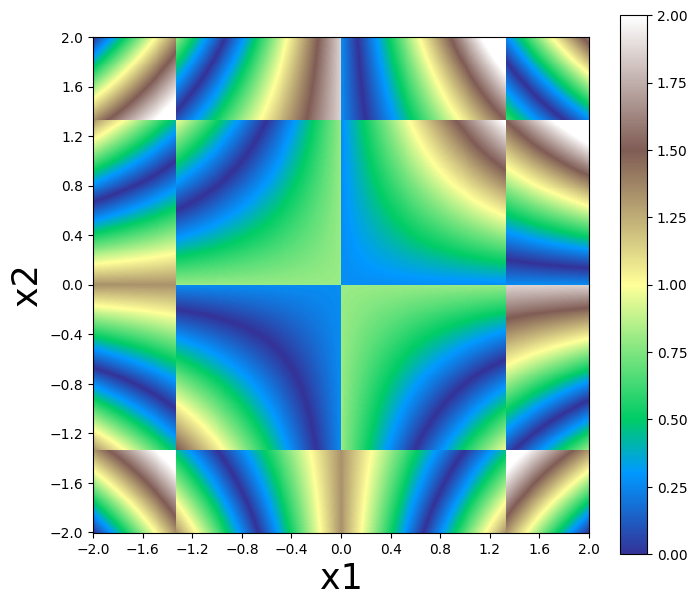}
        \caption{2 bit}
        \label{fig:sub1}
    \end{subfigure}
    \hfill
    \begin{subfigure}[b]{0.245\textwidth}
        \includegraphics[width=\textwidth]{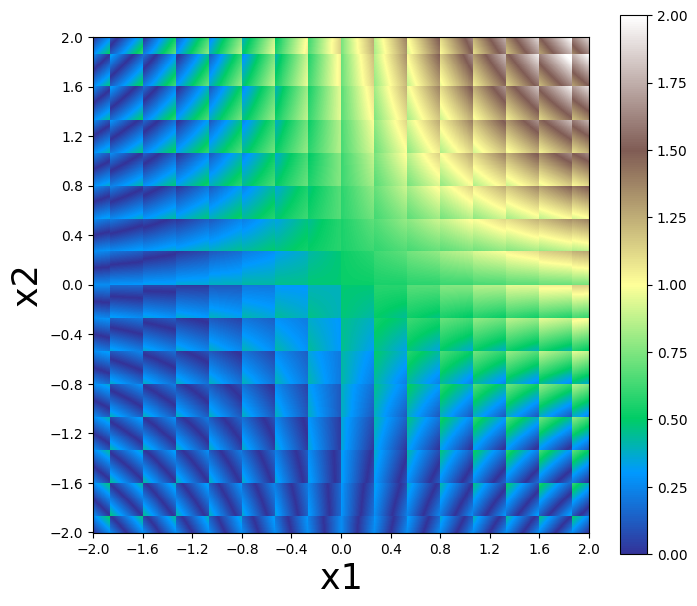}
        \caption{4 bit}
        \label{fig:sub2}
    \end{subfigure}
    \hfill
    \begin{subfigure}[b]{0.245\textwidth}
        \includegraphics[width=\textwidth]{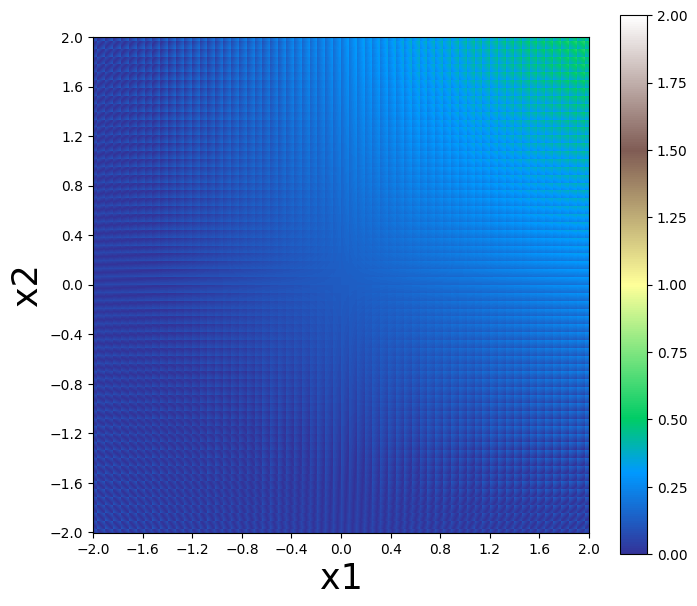}
        \caption{6 bit}
        \label{fig:sub3}
    \end{subfigure}
    \hfill 
    \begin{subfigure}[b]{0.245\textwidth}
        \includegraphics[width=\textwidth]{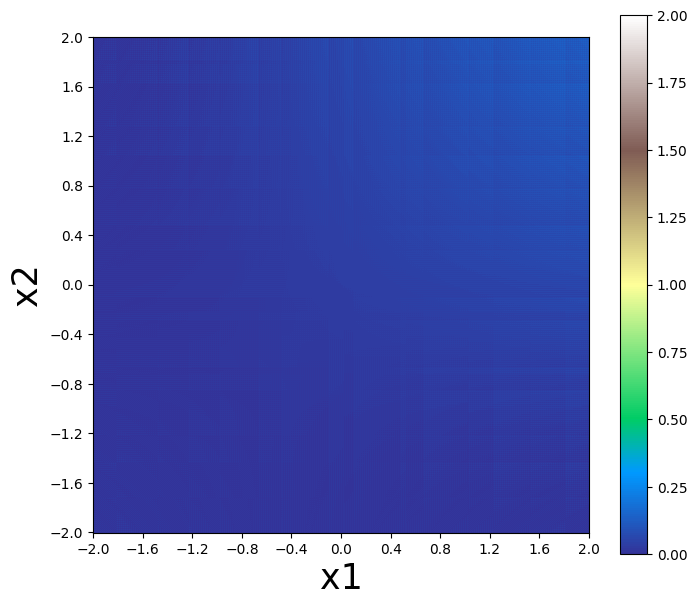}
        \caption{8 bit}
        \label{fig:sub4}
    \end{subfigure}
    \caption{Approximation error $\epsilon_\otimes$ (defined in \Cref{eq:error_prod}) of the DQT integer-only multiplication operator (see \Cref{appendix:A}) compared to standard floating-point multiplication $x_1 \cdot x_2$, evaluated over the interval $[-2, 2]$ for varying bit-widths.  The observed asymmetry around zero arises from the rounding of fixed constants in the DQT implementation; when rounding is removed, the error becomes symmetric.}
    \label{fig:prod_error}
\end{figure*}

\paragraph{Integer Addition Error.}
For the integer addition $k_1 q_1 + k_2 q_2 + k_3$, the total approximation error $\epsilon_{\oplus}$ is a linear combination of the rounding errors, scaled by the integer inputs:
\begin{equation}
    \epsilon_{\oplus} = \delta_1 q_1 + \delta_2 q_2 + \delta_3
\end{equation}
The behavior of this error is illustrated in \Cref{fig:sum_error}.

\paragraph{Integer Multiplication Error.}
Similarly, for integer multiplication, the approximation error $\epsilon_{\otimes}$ is:
\begin{equation}
    \epsilon_{\otimes} = \delta_1 q_1 q_2 + \delta_2 q_1 + \delta_3 q_2 + \delta_4
    \label{eq:error_sum}
\end{equation}
The behavior of this error is illustrated in \Cref{fig:prod_error}.

\paragraph{Integer Dot Product Error.}
In operations such as dot products and convolutions, this error aggregates over the dimension $N$ of the input vectors $\mathbf{x}_q$ and $\mathbf{w}_q$. The total error for the integer dot product from \Cref{eq:integer_dot_product} is:
\begin{equation}
    \epsilon_{\text{dot}} = \delta_1 \sum_{i=1}^N x_{q,i}w_{q,i} + \delta_2 \sum_{i=1}^N x_{q,i} + \delta_3 \sum_{i=1}^N w_{q,i} + \delta_4
    \label{eq:error_prod}
\end{equation}
Although the error terms aggregate over the dimension $N$, the individual rounding errors $\delta_i$ are small, bounded, and have an expected value of zero. In practice, this prevents a catastrophic linear growth of the total error. Furthermore, if the quantized activation and weight distributions are centered around zero (i.e., $\sum x_{q,i} \approx 0$ and $\sum w_{q,i} \approx 0$), the contribution of the $\delta_2$ and $\delta_3$ terms becomes negligible.

\subsection{Error from Integer Division in Bit-Width Transition}
\label{appendix:shift_error}

\begin{figure*}[h!]
    \centering
    \begin{subfigure}[b]{0.245\textwidth}
        \includegraphics[width=\textwidth]{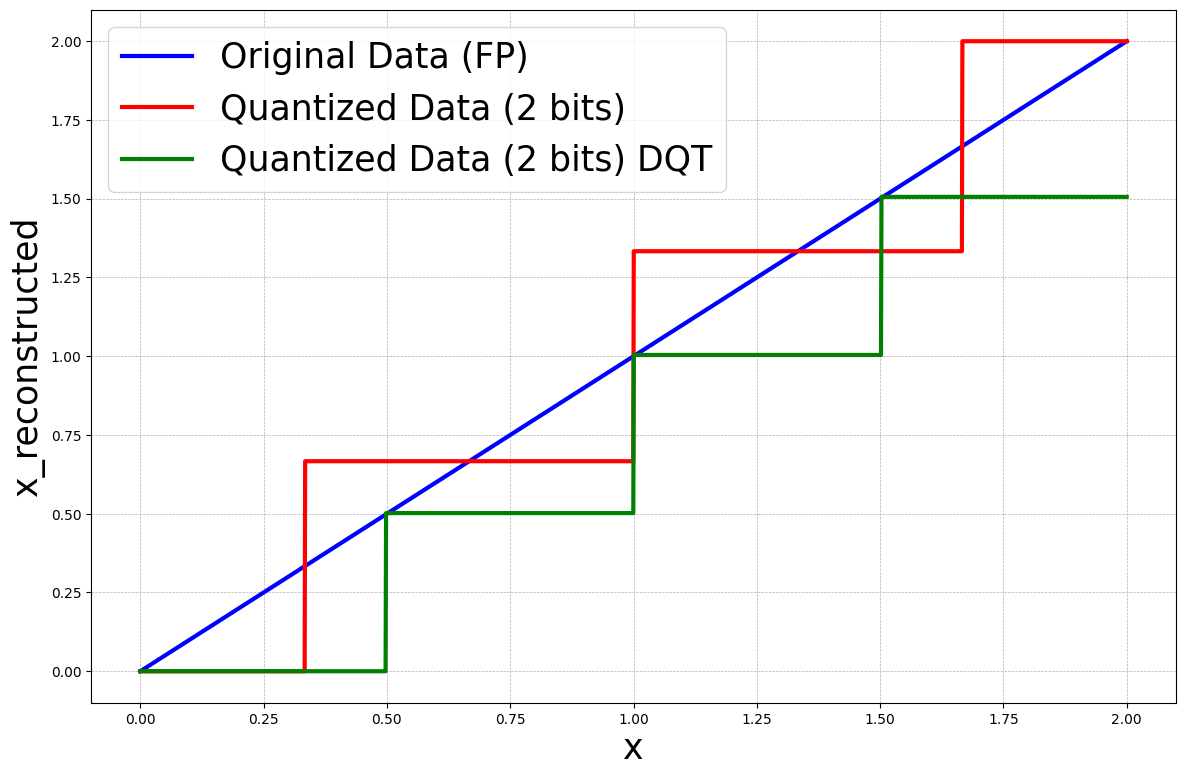}
        \caption{2 bit}
        \label{fig:sub1}
    \end{subfigure}
    \begin{subfigure}[b]{0.245\textwidth}
        \includegraphics[width=\textwidth]{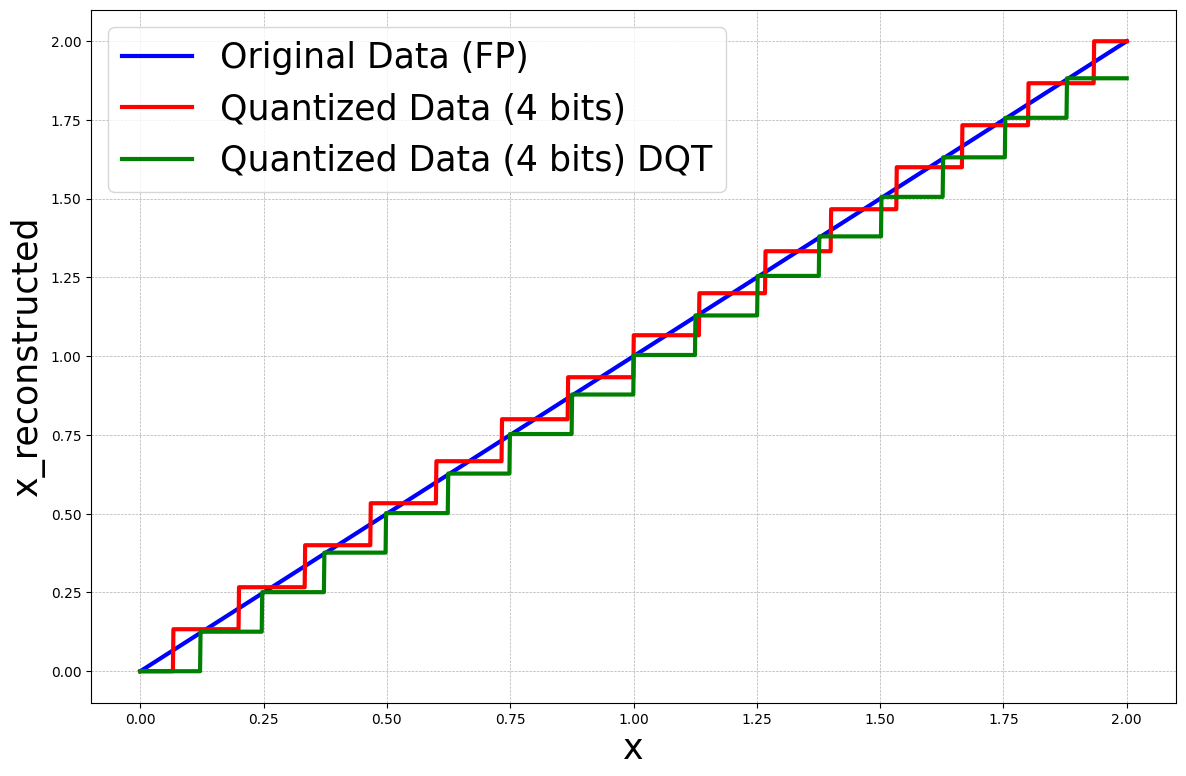}
        \caption{4 bit}
        \label{fig:sub2}
    \end{subfigure}
    \begin{subfigure}[b]{0.245\textwidth}
        \includegraphics[width=\textwidth]{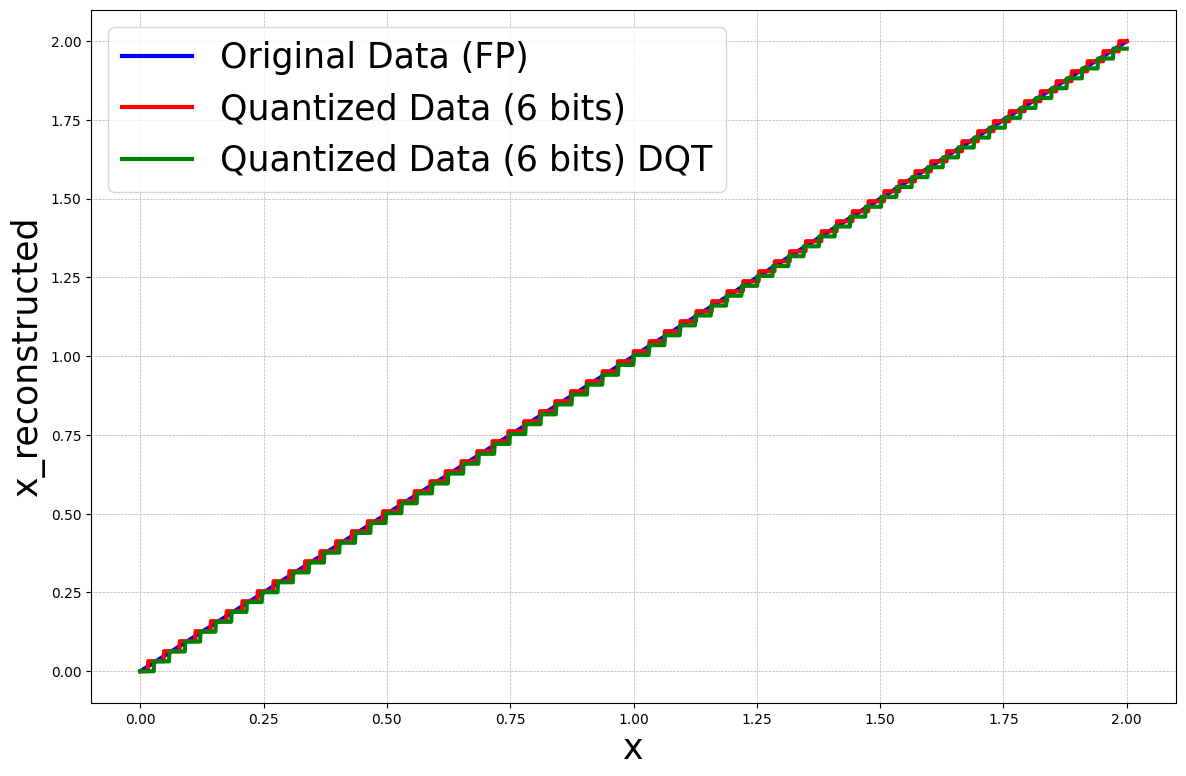}
        \caption{6 bit}
        \label{fig:sub3}
    \end{subfigure}
    \begin{subfigure}[b]{0.245\textwidth}
        \includegraphics[width=\textwidth]{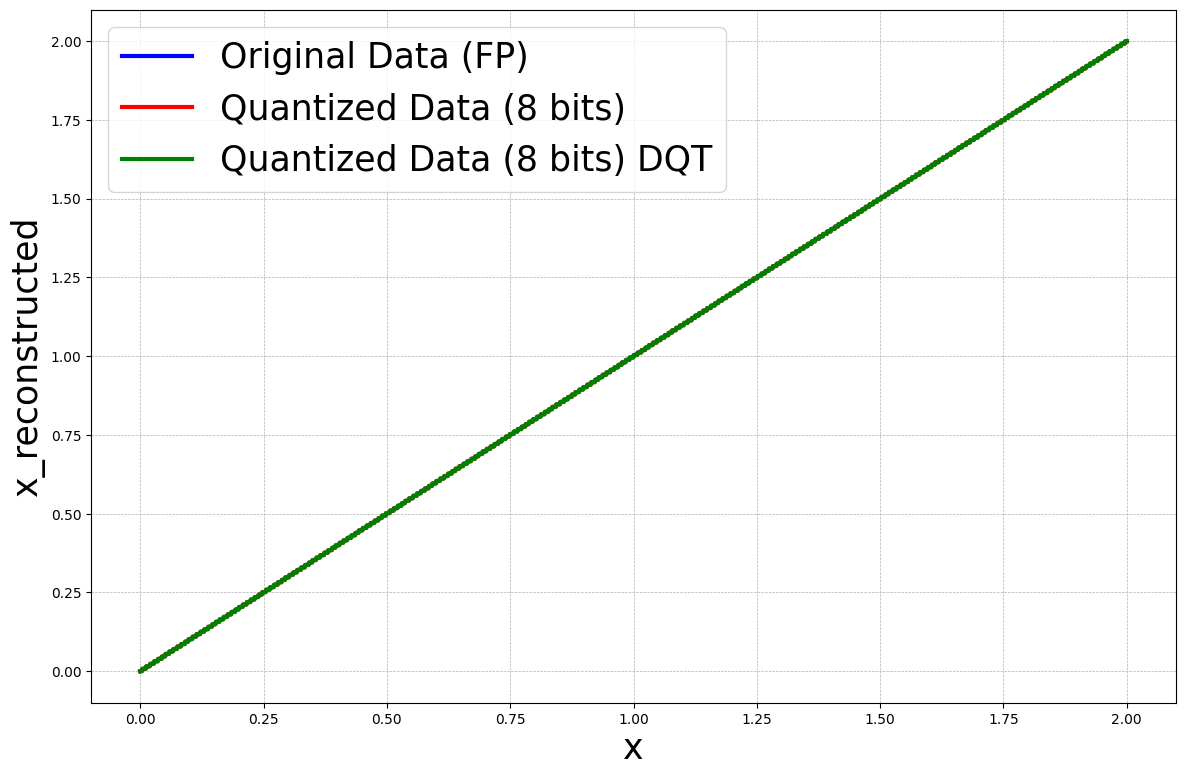}
        \caption{8 bit}
        \label{fig:sub4}
    \end{subfigure}
    \caption{Comparison between our DQT nested quantization scheme (see \cref{eq:bitshift_conversion}) and standard uniform quantization (see \cref{eq:quant}), applied at different bit-widths over the floating-point range \([0, 2]\).}
    \label{fig:error_shift}
\end{figure*}

\begin{figure*}[h!]
    \centering
    \begin{subfigure}[b]{0.245\textwidth}
        \includegraphics[width=\textwidth]{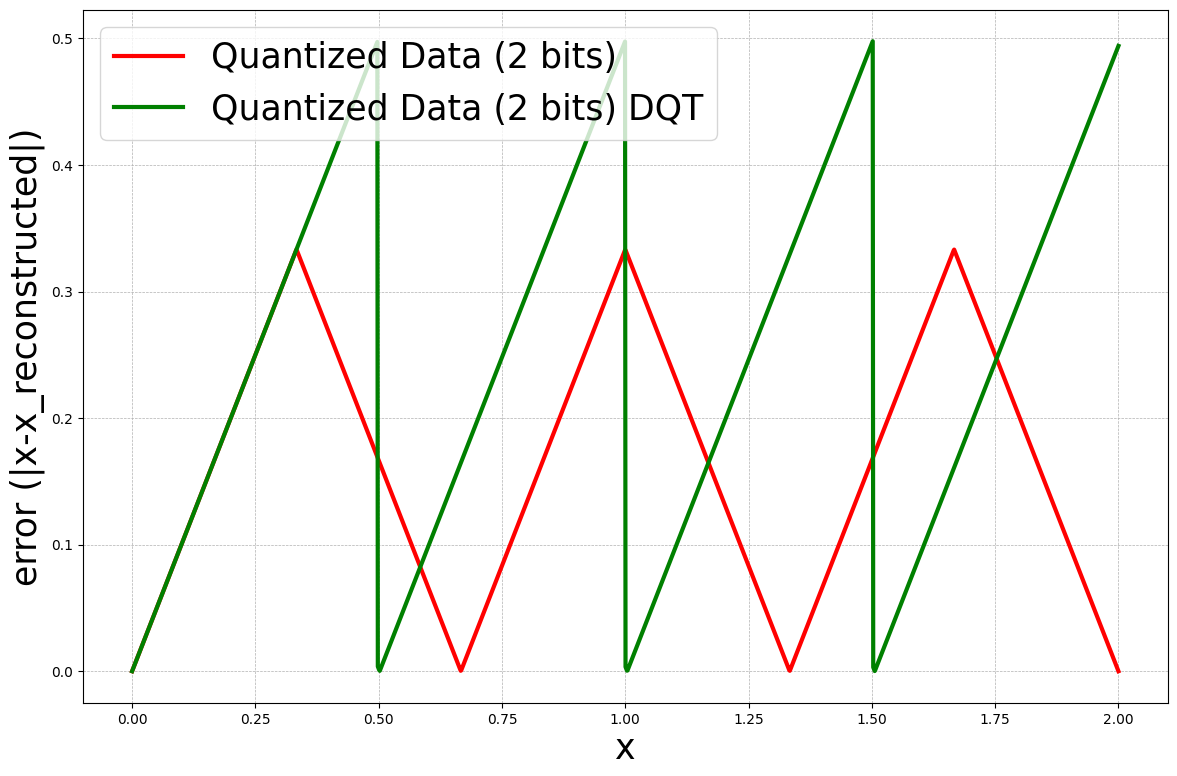}
        \caption{2 bit}
        \label{fig:sub1}
    \end{subfigure}
    \begin{subfigure}[b]{0.245\textwidth}
        \includegraphics[width=\textwidth]{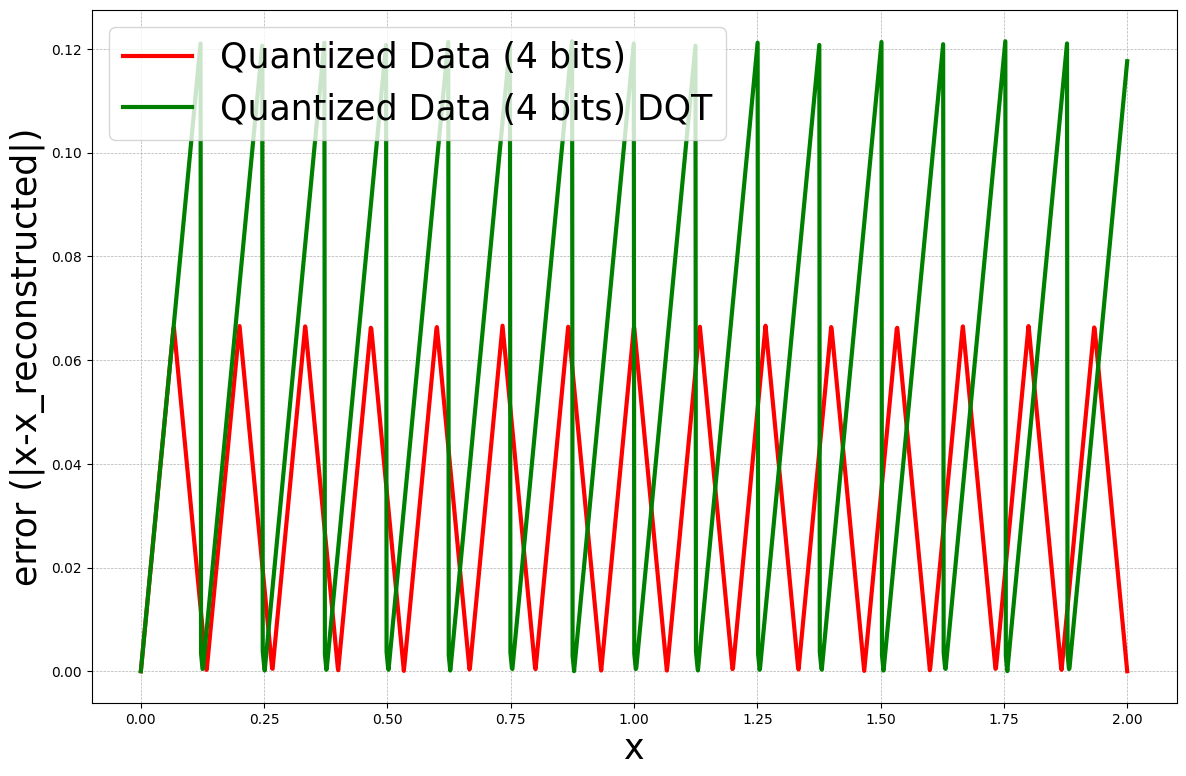}
        \caption{4 bit}
        \label{fig:sub2}
    \end{subfigure}
    \begin{subfigure}[b]{0.245\textwidth}
        \includegraphics[width=\textwidth]{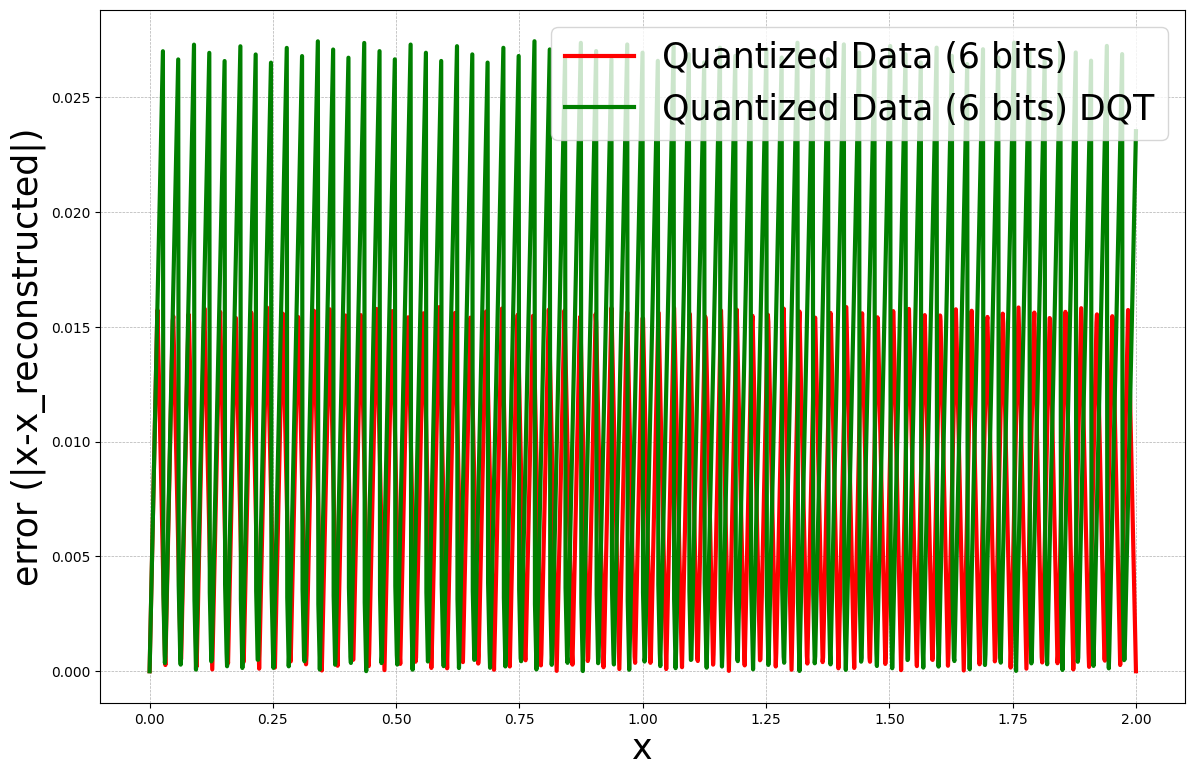}
        \caption{6 bit}
        \label{fig:sub3}
    \end{subfigure}
    \begin{subfigure}[b]{0.245\textwidth}
        \includegraphics[width=\textwidth]{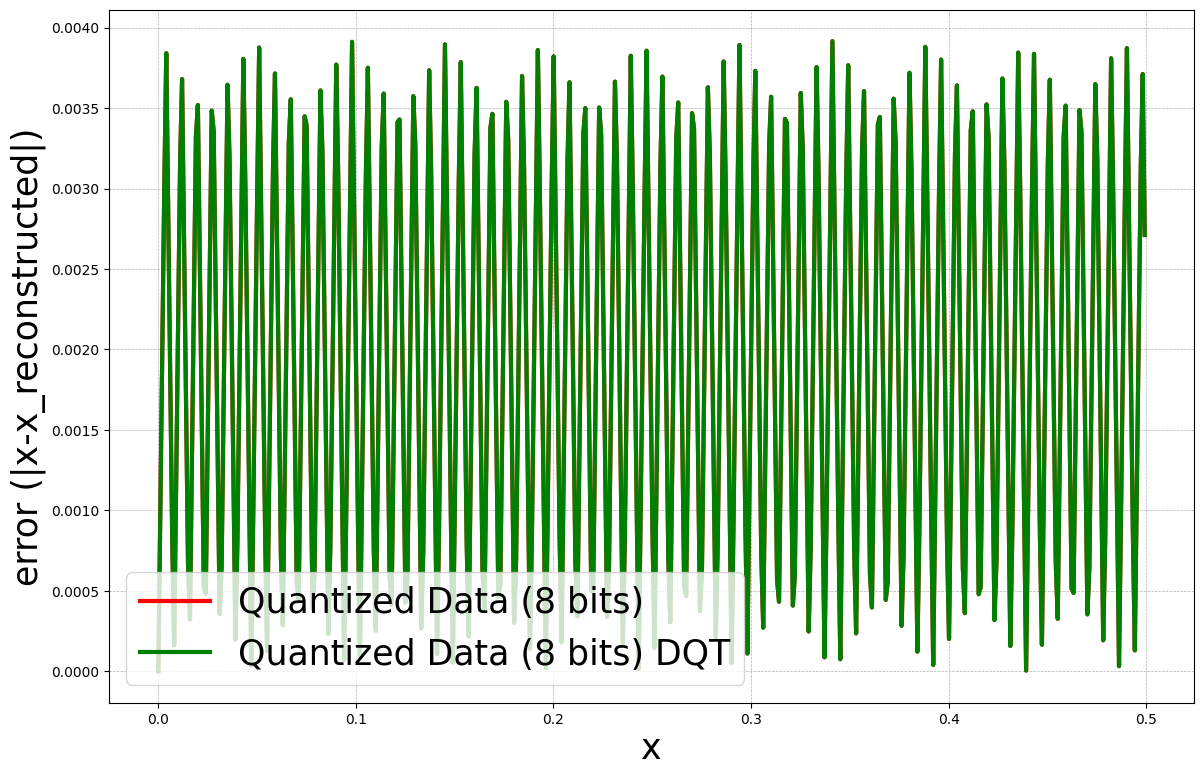}
        \caption{8 bit}
        \label{fig:sub4}
    \end{subfigure}
    \caption{Quantization error for our DQT nested quantization scheme (see \cref{eq:bitshift_conversion}) compared to standard uniform quantization (see \cref{eq:quant}). The error is evaluated over the input range $[0,2]$. For clarity, the 8-bit plot is cropped at an error value of 0.5. Results are shown across multiple bit-widths.
}

    \label{fig:error_shift_2}
\end{figure*}

The primary source of approximation error unique to our framework bit-width transition mechanism arises from the integer division inherent in the process. This section provides a formal analysis of this error.
As derived in Appendix~\ref{appendix:bitshift}, the conversion from a master $n$-bit integer $q_n$ to a target $b$-bit integer $q_b$ is a direct consequence of our power-of-two scale constraint (\Cref{eq:scale_relation}). The computation is~
$
    q_b = \left\lfloor\frac{q_n}{2^{n-b}}\right\rceil
$. 
The exact result of the division $\frac{q_n}{2^{n-b}}$ is a rational number. The approximation error, $\epsilon_\gg$, is introduced by the rounding operation ($\lfloor \cdot \rceil$) that maps this rational number to an integer. The error is formally defined as the difference between the exact value and the rounded result:
\begin{equation}
    \epsilon_\gg = \frac{q_n}{2^{n-b}} - \left\lfloor \frac{q_n}{2^{n-b}} \right\rceil
\end{equation}
This error is inherent to any process that maps a larger discrete range to a smaller one via division and rounding. It is deterministic, systematic, and strictly bounded by $|\epsilon_\gg| \le 0.5$.
The logical right bit-shift operation ($q_n \gg (n-b)$) is a computationally efficient hardware-level implementation of integer division by a power of two (specifically, $\lfloor \frac{q_n}{2^{n-b}} \rfloor$). The rounding-to-nearest-integer operation can be implemented with a small additional integer operation before the shift.
These systematic, bounded errors are illustrated in \Cref{fig:error_shift,fig:error_shift_2}.

\subsection{Mitigation through Quantization-Aware Training}

A key property of our framework is that both sources of error - from operator rounding and bit-width transition - are systematic and deterministic. We integrate both the custom integer-only operators and the bit-shift transition into the forward pass of our network during training. The Straight-Through Estimator (STE) allows gradients to flow through these operations. Consequently, the QAT process is able to learn to adjust the underlying full-precision weights $W$ to compensate for these predictable, quantization-induced discrepancies. 




\section{Experimental Hyperparameters}
\label{appendix:hyperparameters}\label{sec:hyperparameters}

This section details the hyperparameters used for training all models reported in the main paper. All experiments were implemented in PyTorch~\cite{paszke2019pytorchimperativestylehighperformance}.

The following parameters were used for the Stochastic Gradient Descent (SGD) optimizer and associated regularization techniques across all experiments. Optimizer: SGD; momentum: 0.9; batch size: 256; $L_2$ regularization (weight decay): $10^{-5}$.
For, CIFAR-10, models were trained for 200 epochs with an initial learning rate of 0.1 and decay factor of 0.1 applied at epochs 100 and 150.
For ImageNet, models were trained for 120 epochs with an initial learning Rate of 0.01 and cosine annealing schedule.

The following hyperparameters are specific to the DQT framework.
The consistency regularization parameter~$\alpha$ is set to 0.01 for CIFAR-10 and to 0.05 for ImageNet. The cost regularization~$\beta$ is set to 0.0  for the main results to prioritize accuracy. The controller architecture is a lightweight two-layer MLP. The first layer has a hidden dimension of 64. 
The output layer produces a vector of size $N \times M$, where $N$ is the number of layers in the target model, and $M$ is the number of possible bit-width options (e.g., 2, 4, 6, 8 bits). As for the bit-width candidates, we select values centered around the target bit-width. For instance, if the target bit-width is $5$, the candidate set becomes ${4, 5, 6}$, following a strategy similar to that proposed in~\cite{liu2022instance}.
Each element in the output corresponds to a score (or logit) for selecting a specific bit-width for a specific layer. This output can then be passed through a softmax (or Gumbel-Softmax) to sample or select discrete bit-width configurations during training.
\section{Analysis of the Dequantization Bottleneck}
\label{appendix:bottleneck_analysis}
\label{sec:dequantization_bottleneck_analysis}

DQT eliminates the primary performance bottleneck of prior dynamic quantization methods. This section provides a rigorous analysis of this dequantize-requantize cycle and quantifies the performance difference compared to our dequantization-free, bit-shift-based approach.

\subsection{The Standard Floating-Point Cycle}
In conventional dynamic quantization, changing a tensor precision from bit-width $b_1$ to $b_2$ requires routing it through a 32-bit floating-point representation. For each element $q_{b_1}$ in a tensor, with quantization parameters $(\Delta_1, m_1)$ and target parameters $(\Delta_2, m_2)$, the process is:
\begin{itemize}
    \item Dequantization to FP32: $x_\text{fp} = q_{b_1} \cdot \Delta_1 + m_1$
    \item Requantization to INT($b_2$): $q_{b_2} = \text{clip}(\lfloor(x_\text{fp} - m_2) / \Delta_2\rceil, 0, 2^{b_2}-1)$
\end{itemize}
This sequence must be executed for every element of a tensor whose precision is changed at runtime. Its performance cost manifests differently on parallel versus serial architectures.

\begin{table*}[t!]
\centering
\caption{Per-Element Cost Comparison for a Dynamic Precision Change.}
\label{tab:cost_comparison_appendix}
\begin{tabular}{@{}lll@{}}
\toprule
Metric & Standard Dequant-Requant Cycle & Our DQT Bit-Shift \\
\midrule
Operation Type & Arithmetic (Floating-Point) & Logical \\
Data Path & INT $\rightarrow$ FP32 $\rightarrow$ INT & INT $\rightarrow$ INT \\
Primary Cost Driver (CPU) & FP Division Latency & Logical Shift Latency \\
Primary Cost Driver (GPU) & Memory Bandwidth & In-Register Operation \\
Estimated Latency (CPU) & 20--55 cycles & $\sim$1 cycle \\
Estimated Cost (GPU) & High Memory Traffic & Negligible \\
\bottomrule
\end{tabular}
\end{table*}
\subsection{Cost Analysis on Parallel Architectures (GPUs)}
On a GPU, performance is primarily constrained by memory bandwidth. The cost of the dequantize-requantize cycle is not the latency of individual arithmetic operations, but the substantial memory traffic it generates. The process requires reading the original $b_1$-bit integer tensor from global memory, dequantizing it, and writing a new, temporary FP32 tensor back to global memory. This intermediate tensor is $32/b_1$ times larger than the original, creating a significant memory traffic overhead. This large temporary tensor must then be read from memory again to perform the requantization. This memory-bound process, combined with the overhead of launching multiple compute kernels, saturates the memory bus and breaks the optimized integer-only dataflow of modern accelerators.

\subsection{Cost Analysis on Serial Architectures (CPUs)}
On a CPU, performance is primarily dictated by instruction latency, measured in clock cycles. The dequantize-requantize cycle is latency-bound due to its sequence of floating-point arithmetic operations. For each element, the critical path includes floating-point multiplication, addition/subtraction, and division. While multiplication and addition are fast (typically 3-5 cycles on modern architectures), floating-point division is a high-latency, non-pipelined operation. Its latency can range from 10 to over 40 clock cycles, making it the dominant cost. The total latency for the sequence, including data type conversions, can exceed 20-55 cycles per element, rendering it prohibitively expensive for large tensors.

\subsection{The DQT Alternative: A Single Logical Operation}
In contrast, our DQT framework performs the same precision change with a single, low-latency logical operation:
\begin{equation}
    q_{b_2} = q_{b_1} \gg (b_1 - b_2)
\end{equation}
This operation is executed by a single-cycle circuit (a barrel shifter) within the processor Arithmetic Logic Unit (ALU). On a GPU, this completely eliminates the creation of the intermediate FP32 tensor and the associated memory traffic, as the conversion can be performed in-register within a single compute kernel. On a CPU, it replaces the $\sim$20-55 cycle arithmetic sequence with a $\sim$1 cycle logical instruction. In both cases, the result is a performance improvement of one to two orders of magnitude for the bit-width transition step, enabling truly efficient dynamic inference.

\subsection{Quantitative Cost Comparison}
The architectural differences between the standard dequantize-requantize cycle and our bit-shift mechanism result in a performance gap of one to two orders of magnitude for the bit-width transition step. Table~\ref{tab:cost_comparison_appendix} provides a comparative analysis of the per-element cost, summarizing the principles discussed in this section. As shown in our main results (Table~\ref{tab:results_main}), this fundamental efficiency advantage allows DQT to replace millions of high-latency floating-point operations required by prior work with an equivalent number of low-cost logical shifts.

\section{Analysis of Regularization Hyperparameters}
\label{sec:appendix_hyperparams}
\label{appendix:alpha_beta_effect}

The DQT objective introduced in the main paper includes two regularization hyperparameters, $\alpha$ and $\beta$, that control consistency and cost:
\begin{equation}
    J = J_{\text{task}} + \alpha \cdot J_{\text{consistency}} + \beta \cdot J_{\text{cost}}
\end{equation}

This section provides an empirical analysis of how these hyperparameters affect training stability and the trade-off between accuracy and efficiency. All experiments were conducted using ResNet-18 on CIFAR-10. For the analysis in this section, the controller was trained with a fixed set of candidate bit-widths: \{2, 4, 8\} bits.

\subsection{Effect of the Consistency Hyperparameter \(\alpha\)}

\begin{figure}[h!]
    \centering
    \begin{subfigure}{0.45\textwidth}
        \includegraphics[width=\linewidth]{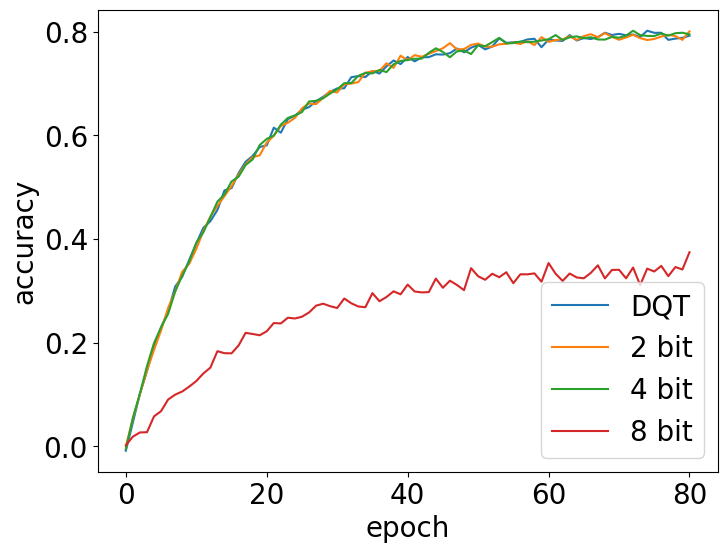}
        \caption{\(\alpha = 0\): Consistency loss disabled}
        \label{fig:alpha_zero}
    \end{subfigure}
    \hfill
    \begin{subfigure}{0.45\textwidth}
        \includegraphics[width=\linewidth]{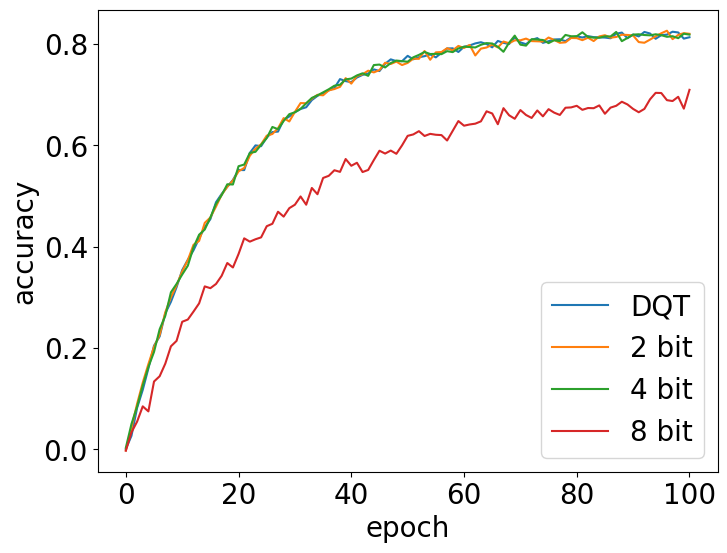}
        \caption{\(\alpha = 0.05\): Consistency loss enabled}
        \label{fig:alpha_nonzero}
    \end{subfigure}
    \caption{
        Impact of the consistency hyperparameter $\alpha$.  
        (a) Without consistency loss, low-bit configurations converge poorly.  
        (b) With consistency loss, all bit-widths perform well, indicating more stable and generalizable training.
    }
    \label{fig:alpha_effect}
\end{figure}

The consistency loss term, \(J_{\text{consistency}}\), encourages the shared backbone network to generalize well across all candidate bit widths, even those less frequently selected by the controller. This helps stabilize training and improve robustness.
To isolate the impact of $\alpha$, we fix $\beta = 0$ and compare two settings, one in which the consistency loss is disabled~($\alpha = 0$), and on in which the consistency loss is enabled~($\alpha = 0.05$).
\Cref{fig:alpha_effect} shows validation accuracy over 100 epochs for both the full DQT model and static models with fixed 2-, 4-, and 8-bit configurations.

The results indicate that when~$\alpha = 0$, training is unstable for the 2-bit configuration (\Cref{fig:alpha_zero}). These paths receive fewer useful gradient updates, as the controller favors higher-precision bit-widths during training. When~$\alpha = 0.05$ instead, all configurations achieve strong performance (\Cref{fig:alpha_nonzero}). The consistency loss forces the model to perform well across all bit-widths, leading to better generalization and true dynamic behavior.
Setting $\alpha > 0$ is essential to maintain the effectiveness of the shared network weights for various candidate bit-widths. This ensures a training process that is stable and results in a model capable of handling precision variations during inference. Crucially, it supports \textit{deployment-time flexibility}: the model continues to function reliably even with limited bit-width options (e.g., $\{2, 4\}$ instead of the full set $\{2, 4, 8\}$) caused by hardware or energy limitations.

\subsection{Effect of the Cost Hyperparameter \(\beta\)}

The cost loss \(J_{\text{cost}}\) penalizes high-precision bit-widths, allowing explicit control over the model’s average bit usage and computational footprint.

\begin{figure}[h!]
    \centering
    \includegraphics[width=\linewidth]{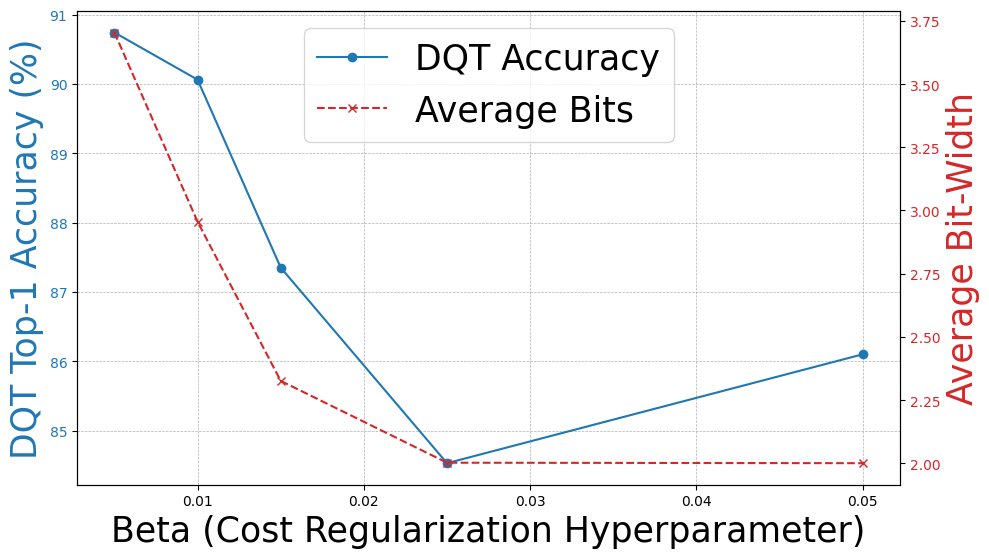}
    \caption{
        Trade-off between accuracy and average bit-width as \(\beta\) increases.  
        Larger \(\beta\) values lead to more efficient (lower-bit) models, with gradually reduced accuracy.
    }
    \label{fig:beta_effect}
\end{figure}

To analyze the effect of $\beta$, we fix $\alpha = 0$ and vary $\beta$ in the range $[0.005,0.05]$. For each setting, we measure Top-1 validation accuracy and the average bit-width selected by the controller.
\Cref{fig:beta_effect} summarize the results.

The results shows that higher $\beta$ values penalize the use of higher-precision operations, resulting in a monotonic decrease in the average bit-width selected. This results in a smooth, controlled decline in model accuracy, allowing practitioners to find an optimal balance for a given resource budget. Low values of the cost hyperparameter $\beta$ maximize accuracy, while higher values promote lightweight models suitable for deployment on constrained hardware.

\section{Computational Cost Analysis of DQT's Integer MAC}
\label{appendix:dqt_cost_analysis}

This appendix provides a detailed analysis of the primitive integer operations required for a single Multiply-Accumulate (MAC) step within our DQT framework. We contrast our formulation with a conventional quantized MAC to precisely quantify the computational trade-offs.

\begin{table*}[t!]
\centering
\caption{Per-Element Primitive Operation Cost of the Inner MAC Loop.}
\label{tab:cost_comparison_final}
\begin{tabular}{@{}lcc@{}}
\toprule
Primitive Operation & Standard Integer MAC & Optimized DQT MAC \\
\midrule
Integer Multiplication & 1 & 1 \\
Integer Addition/Subtraction & 3 & 2 \\
\bottomrule
\end{tabular}
\end{table*}

\subsection{Conventional Integer MAC}
In a standard integer-only inference pipeline, the dot product between an activation vector $\mathbf{x}_q$ and a weight vector $\mathbf{w}_q$ is typically computed as:
\begin{equation}
\label{eq:standard_mac_sum}
    \text{acc}_\text{int32} = \sum_{i=1}^N (x_{q,i} - z_x)(w_{q,i} - z_w)
\end{equation}
where $z_x$ and $z_w$ are integer zero-points. The high-precision integer accumulator is then rescaled to the output precision using a single fixed-point multiplication (requantization) after the summation is complete. The inner loop, which dominates the computation, requires one multiplication and three additions/subtractions per element.

\subsection{DQT Integer MAC}
In our framework, the MAC is implemented using the integer dot product formulation from \Cref{eq:integer_dot_product}. The key difference is that scaling and offset corrections are integrated into the accumulation loop.

\paragraph{General Formulation.}
In the general case with non-zero offsets ($m_x, m_w \ne 0$), the contribution of each element to the final sum is:
\begin{equation}
\label{eq:dqt_mac_general}
    \text{term}_i = k_1 x_{q,i}w_{q,i} + k_2 x_{q,i} + k_3 w_{q,i}
\end{equation}
A naive implementation would compute this term for each element and accumulate the results, requiring three multiplications and two additions per element within the loop.

\paragraph{Optimized Formulation with PACT and Reordering.}
In our final DQT implementation, we make two critical optimizations. First, we use a PACT-like scheme for activations, which sets their range to $[0, \alpha]$, resulting in $m_x=0$. This simplifies the dot product by eliminating the $k_3$ term, as its numerator $\Delta_w m_x$ becomes zero. Second, we reorder the computation by factoring the constant multipliers $k_1$ and $k_2$ out of the summation:
\begin{equation}
\label{eq:dqt_mac_optimized}
\begin{aligned}
    q_{\text{dot}} &= \sum_{i=1}^N (k_1 x_{q,i} w_{q,i} + k_2 x_{q,i}) + k_4' \\
    &= k_1 \left( \sum_{i=1}^N x_{q,i} w_{q,i} \right) + k_2 \left( \sum_{i=1}^N x_{q,i} \right) + k_4'
\end{aligned}
\end{equation}
where $k_4' = k_4$ from the dot product derivation. This reordering fundamentally changes the cost structure. The expensive inner loop is reduced to accumulating two sums: $\sum x_{q,i} w_{q,i}$ and $\sum x_{q,i}$. This requires only one multiplication and two additions per element. The scaling by $k_1$ and $k_2$ is performed only once after the loop is complete.

\subsection{Summary of Per-Element Costs}
Table~\ref{tab:cost_comparison_final} summarizes the per-element operational cost of the inner loop for each MAC formulation. Our optimized DQT MAC achieves a cost nearly identical to the standard integer MAC. However, by integrating scaling into its formulation, it avoids the final requantization step that relies on floating-point parameters, thereby enabling a fully dequantization-free pipeline suitable for dynamic precision changes. This design choice is what allows DQT to eliminate the system-level bottleneck of the dequantize-requantize cycle.

\end{document}